\documentclass[10pt,twocolumn,letterpaper]{article}

\usepackage[pagenumbers]{cvpr} 

\newcommand{\method}{SAEF\xspace}
\usepackage{subcaption}
\usepackage{amssymb}
\usepackage{amsmath}
\usepackage{multirow}

\usepackage{xspace}
\usepackage{enumitem}
\usepackage{graphicx}  
\usepackage{caption} 
\usepackage{amsmath}
\usepackage{wrapfig}
\usepackage{multirow}
\usepackage{colortbl}
\usepackage{pifont}
\usepackage{algorithm}
\usepackage{algorithmicx}
\usepackage[noend]{algpseudocode}
\definecolor{cvprblue}{rgb}{0.21,0.49,0.74}
\usepackage[pagebackref,breaklinks,colorlinks,allcolors=cvprblue]{hyperref}

\def\paperID{} 
\def\confName{CVPR}
\def\confYear{2026}

\title{From Isolation to Integration: Building an Adaptive Expert Forest for Pre-Trained Model-based Class-Incremental Learning}

\author{
  Ruiqi Liu\text{$^{1,2}$}, Boyu Diao\text{$^{1,2,\dagger}$}, Hangda Liu\text{$^{1,2}$}, Zhulin An\text{$^{1,2}$}, \\
  Fei Wang\text{$^{1,2}$}, Yongjun Xu\text{$^{1,2}$}\\
  $^{1}$Institute of Computing Technology, Chinese Academy of Sciences, Beijing, China\\
  $^{2}$University of Chinese Academy of Sciences, Beijing, China\\
  \texttt{liuruiqi23@mails.ucas.ac.cn},\\
  \texttt{\{diaoboyu2012,	liuhangda21s, wangfei, anzhulin, xyj\}@ict.ac.cn}, \\
}

\begin{document}
\maketitle
\begin{abstract}
Class-Incremental Learning (CIL) requires models to learn new classes without forgetting old ones. A common method is to freeze a pre-trained model and train a new, lightweight adapter for each task. While this prevents forgetting, it treats the learned knowledge as a simple, unstructured collection and fails to use the relationships between tasks. To this end, we propose the Semantic-guided Adaptive Expert Forest (SAEF), a new method that organizes adapters into a structured hierarchy for better knowledge sharing. SAEF first groups tasks into conceptual clusters based on their semantic relationships. Then, within each cluster, it builds a balanced expert tree by creating new adapters from merging the adapters of similar tasks. At inference time, SAEF finds and activates a set of relevant experts from the forest for any given input. The final prediction is made by combining the outputs of these activated experts, weighted by how confident each expert is. Experiments on several benchmark datasets show that SAEF achieves SOTA performance.
\end{abstract}

\begin{figure}[t!]
    \centering
    \includegraphics[width=\columnwidth]{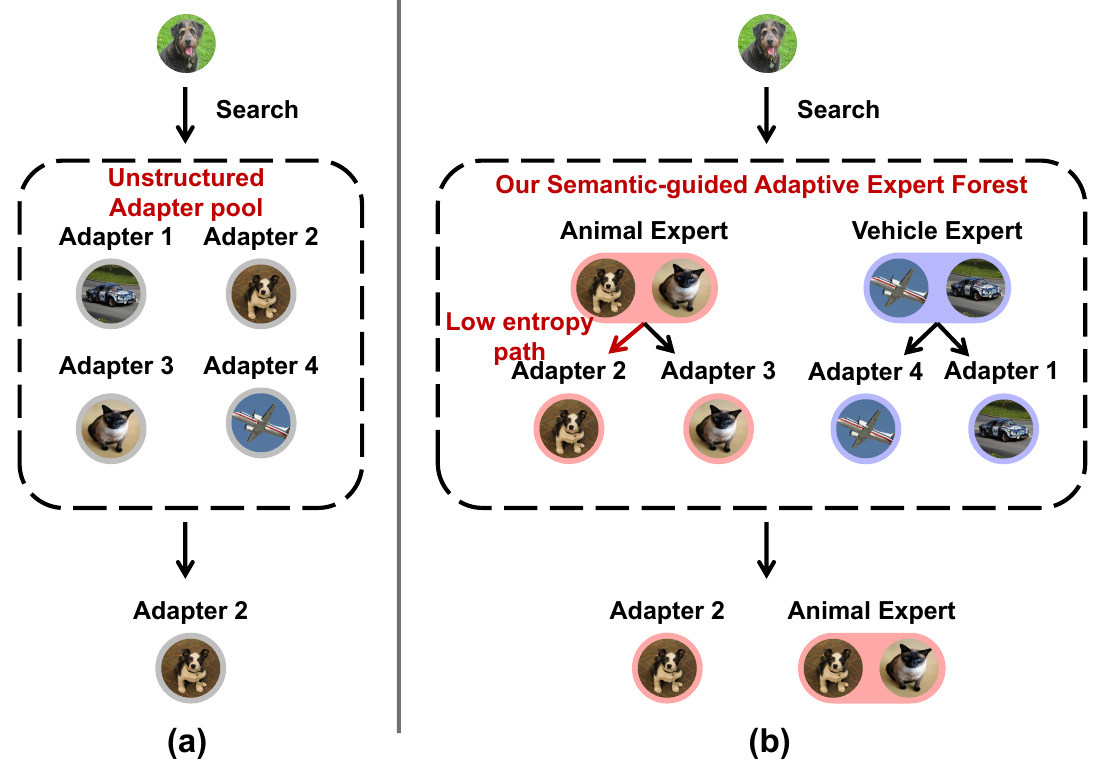}
    \caption{
        Conceptual comparison of adapter-based CIL methods.
        \textbf{(a)} The conventional methods treat adapters as an unstructured pool, leading to inefficient full-ensemble inference. 
        \textbf{(b)} Our proposed SAEF automatically organizes adapters into a semantic expert forest, enabling an efficient, adaptive search for the most relevant experts.
    }
    \label{fig:intro}
\end{figure}

\section{Introduction}
\label{sec:intro}
Class-Incremental Learning (CIL) enables models to continually acquire knowledge of new classes while preserving performance on previously learned ones~\cite{mccloskey1989catastrophic, kirkpatrick2017overcoming}. While mitigating catastrophic forgetting is a primary objective, the growing application of CIL in resource-sensitive contexts, such as on-device and robotic systems, has also underscored the importance of computational efficiency~\cite{soltoggio2024collective, liu2024resource}.
CIL methodologies can be broadly categorized into two paradigms: training from scratch and leveraging Pre-trained Models (PTMs). The former category, which includes regularization, architectural, and rehearsal-based methods~\cite{li2017learning, serra2018overcoming, rebuffi2017learning}, often encounters practical limitations concerning scalability, data privacy, and overall performance~\cite{wang2022dualprompt}. Consequently, a significant portion of recent research has shifted towards PTM-based CIL~\cite{zhou2024revisiting}. This method utilizes the powerful representations learned during large-scale pre-training as a robust foundation for incremental tasks, frequently leading to state-of-the-art (SOTA) results. A prominent strategy within this paradigm involves freezing the PTM backbone and applying Parameter-Efficient Fine-Tuning (PEFT) techniques, such as training lightweight adapters~\cite{sun2025mos} or prompts~\cite{wang2022learning, smith2023coda}. This method offers a compelling balance of high performance and low computational overhead, making it a highly influential and widely adopted method in the field.

A common strategy within this PTM-based paradigm is to train a separate, lightweight adapter for each incremental task~\cite{zhou2024expandable}. During inference, systems may either retrieve the most suitable task adapter~\cite{smith2023coda} or ensemble predictions from the entire collection~\cite{zhou2024expandable}. While separating task-specific parameters helps prevent catastrophic forgetting, this method treats the learned knowledge as a simple, unstructured collection, as shown in Fig.~\ref{fig:intro}(a). This design fails to use the important semantic relationships between tasks (for example, between a 'dog' task and a 'cat' task) for better knowledge transfer. Some methods attempt to address this by adding a global module to learn shared knowledge from all tasks~\cite{huai2025cl}. However, we argue that sharing knowledge globally across all tasks is a suboptimal method. True knowledge transfer should be more selective and structured. It should be context-specific and hierarchical, sharing information only between related tasks, as illustrated in Fig.~\ref{fig:intro}(b).

To this end, we propose the Semantic-guided Adaptive Expert Forest (SAEF). SAEF is a method that automatically organizes the set of task-specific adapters into a structured expert forest for better knowledge sharing. This process consists of three main stages. First, SAEF groups tasks into conceptual clusters based on their semantic relationships. Second, within each cluster, it constructs a balanced expert tree by recursively creating new adapters from merging the adapters of the most similar tasks. This process builds a hierarchy of experts at different levels of abstraction. Finally, the third stage involves adaptive inference, where SAEF employs a parallel search strategy across this hierarchy to dynamically activate a subset of relevant experts for any given sample. The final prediction is produced by an entropy-guided fusion of these experts' outputs. In summary, our main contributions are as follows:
\begin{itemize}
\item We propose SAEF, a novel method that automatically constructs a semantic expert hierarchy from previously isolated task adapters to enable structured knowledge reuse.
\item We design a dynamic inference strategy that adaptively navigates this hierarchy, using a multi-path search and an entropy-guided fusion mechanism to ensure robust predictions.
\item We conduct extensive experiments on major CIL benchmarks, where SAEF consistently outperforms SOTA methods with accuracy gains of up to 1.34\%.
\end{itemize}
\section{Related Work}

CIL methods can be broadly divided into two main categories: training models from scratch and using Pre-trained Models (PTMs).

\subsection{Training Models from Scratch}
The traditional method for CIL is to train a model from scratch. These methods are typically grouped into three categories. \textbf{Regularization-based} methods~\cite{yu2020semantic,serra2018overcoming,li2017learning,yang2024clip,chaudhry2018riemannian,aljundi2018memory} add a loss term to penalize changes to important parameters from past tasks. \textbf{Architecture-based} methods~\cite{ebrahimi2020adversarial,ke2020continual,loo2020generalized,wang2020learn,zhao2022deep} add new parameters for each new task to keep knowledge separate, for example, by growing the network or using different parts of it for different tasks. \textbf{Rehearsal-based} methods~\cite{aljundi2019gradient,buzzega2020dark,cha2021co2l,chaudhry2021using,chaudhry2018efficient} store a small number of old samples and replay them during training on new tasks. These methods have limitations in scalability and performance, which has led to a shift towards PTM-based methods~\cite{buzzega2020dark, zhu2021class, wang2022foster, shokri2015privacy, smith2021always, wang2022dualprompt, li2024fcs}.

\subsection{Pre-Trained Model-based CIL}
To avoid the high cost of training from scratch, modern CIL often relies on a frozen pre-trained model (PTM) with Parameter-Efficient Fine-Tuning (PEFT). This involves training only a small set of new parameters for each task. This makes CIL practical for many resource-constrained applications. Common PEFT methods include prompt tuning, which learns to select or combine prompts~\cite{wang2022learning, wang2022dualprompt, smith2023coda}, and adapter-based tuning, which adds a new adapter for each task~\cite{sun2025mos, zhou2024expandable}. Other methods also follow this paradigm, such as fine-tuning only the final layer~\cite{zhang2023slca} or using random projections~\cite{mcdonnell2024ranpac}.

Recently, the focus has shifted towards enabling knowledge sharing between these task-specific modules. For instance, some methods learn to route inputs using a Mixture-of-Experts (MoE) method~\cite{fan2022m3vit,hazimeh2021dselect,ho2022convergence,jacobs1991adaptive}, while others create a shared global expert~\cite{sun2025mos}. A key limitation of these methods, however, is that they treat the set of experts or adapters as an unstructured collection. This overlooks the underlying semantic hierarchy of tasks, where some tasks are inherently more related than others. Our work addresses this exact limitation by proposing a method that organizes adapters based on their semantic relationships, enabling more structured and effective knowledge sharing.

\section{Methodology}
\label{sec:methodology}

\subsection{Preliminaries}

\paragraph{Class-Incremental Learning (CIL).}
In Class-Incremental Learning, a model learns from a sequence of tasks, where each new task introduces a new set of classes. We consider a sequence of $T$ tasks, $\{\mathcal{D}_1, \mathcal{D}_2, \ldots, \mathcal{D}_T\}$. For each task $t$, the model is given a training set $\mathcal{D}_t = \{(x_i, y_i)\}_{i=1}^{N_t}$, where $x_i$ is an image and $y_i$ is its class label from the set $\mathcal{C}_t$. The sets of classes are disjoint for different tasks, so $\mathcal{C}_t \cap \mathcal{C}_{t'} = \emptyset$ for all $t \neq t'$.

During the training phase for task $t$, the model only has access to the data in $\mathcal{D}_t$. Our work follows the challenging exemplar-free setting~\cite{wang2022learning,wang2022dualprompt,zhou2024revisiting}, meaning no samples from past tasks are stored for replay. The model consists of a feature extractor $\phi$ and a classifier $W$. The set of all trainable parameters for task $t$ is denoted as $\theta_t$. After learning from task $t$, the model's performance is evaluated on all classes seen so far, which are in the set $\mathcal{Y}_t = \bigcup_{i=1}^{t} \mathcal{C}_i$. The main goal is to find the optimal parameters $\theta_t^*$ that minimize the loss over the test sets of all tasks:
\begin{equation}
\label{eq:cil_objective}
\theta_t^* = \underset{\theta_t}{\operatorname{argmin}} \sum_{i=1}^{t} \mathbb{E}_{(x,y) \sim \mathcal{D}^{\text{test}}_i} \left[ \mathcal{L}_{\text{CE}}(W^\top \phi(x; \theta_t), y) \right],
\end{equation}
where $W^\top \phi(x; \theta_t)$ are the output logits from the model, $\mathcal{D}^{\text{test}}_i$ is the test set for the $i$-th task, and $\mathcal{L}_{\text{CE}}$ is the standard cross-entropy loss. A successful CIL model needs to be plastic enough to learn new classes from $\mathcal{D}_t$ while remaining stable to avoid forgetting the old classes in $\mathcal{Y}_{t-1}$.

\paragraph{PTM-based CIL with Task-Specific Adapters.}
Our method is based on the modern PTM-based CIL framework. In this framework, a large pre-trained model, like a ViT~\cite{dosovitskiy2020image}, is used as a fixed backbone. For each new task, only a small, lightweight adapter module is trained. This partitions the trainable parameters $\theta_t$ into a set of task-specific adapters, $\mathcal{A}_t$, and a classifier that is updated for new classes.

Following standard practice~\cite{zhou2024revisiting,sun2025mos}, adapters are small bottleneck modules inserted into each Transformer block of the PTM. An adapter is usually added in parallel to the MLP layer. For an input feature $\mathbf{x}_i \in \mathbb{R}^d$ to the MLP, the forward pass with an adapter is computed as:
\begin{align} \label{eq:adapter_forward}
	\mathbf{x}_o = \text{MLP}(\mathbf{x}_i) + \text{ReLU}(\mathbf{x}_i W_{\text{down}})W_{\text{up}},
\end{align}
where $\mathbf{x}_o \in \mathbb{R}^d$ is the output. The adapter consists of a down-projection matrix $W_{\text{down}} \in \mathbb{R}^{d \times r}$ and an up-projection matrix $W_{\text{up}} \in \mathbb{R}^{r \times d}$, where the bottleneck dimension $r$ is much smaller than $d$. For each new task $t$, a new adapter $\mathcal{A}_t = \{W_{\text{down}}^{(l,t)}, W_{\text{up}}^{(l,t)}\}_{l=1}^L$ is initialized and trained.

To reduce interference between different task adapters, the training process for $\mathcal{A}_t$ on task $\mathcal{D}_t$ combines two loss functions. The first is the standard cross-entropy classification loss, $\mathcal{L}_{\text{cls}}$, which is computed over all classes seen so far, $\mathcal{Y}_t$:
\begin{equation}
\label{eq:cls_loss}
    \mathcal{L}_{\text{cls}} = -\frac{1}{N_t} \sum_{(\mathbf{x}, y) \in \mathcal{D}_t} \log \frac{\exp(\mathbf{w}_y^\top\phi(\mathbf{x};\mathcal{A}_t))}{\sum_{c \in \mathcal{Y}_{t}} \exp(\mathbf{w}_c^\top\phi(\mathbf{x};\mathcal{A}_t))}.
\end{equation}
Here, $\phi(\mathbf{x};\mathcal{A}_t)$ is the feature vector produced by the model with the new adapter $\mathcal{A}_t$, and $\mathbf{w}_y$ is the classifier weight for class $y$. The second loss is an orthogonality regularization term, $\mathcal{L}_{\text{orth}}$, which encourages the up-projection matrix of the new adapter to be orthogonal to those of all previous adapters:
\begin{align} \label{eq:adapter-orth}
    \mathcal{L}_{\text{orth}} = \sum_{l=1}^{L} \sum_{i=1}^{t-1}\left\| {W_{\text{up}}^{(l,t)}} \cdot  {(W_{\text{up}}^{(l,i)})}^\top \right\|_{F},
\end{align}
where $W_{\text{up}}^{(l,t)}$ is the up-projection matrix of the adapter in the $l$-th layer for the current task $t$, and $\| \cdot \|_{F}$ denotes the Frobenius norm. The overall loss for training the adapter $\mathcal{A}_t$ is a weighted sum of these two components:
\begin{align} \label{eq:adapter-overall}
    \mathcal{L} = \mathcal{L}_{\text{cls}} + \lambda \mathcal{L}_{\text{orth}},
\end{align}
where $\lambda$ is a hyperparameter that balances the two objectives.

\begin{figure*}[t!]
    \centering
    \includegraphics[width=\textwidth]{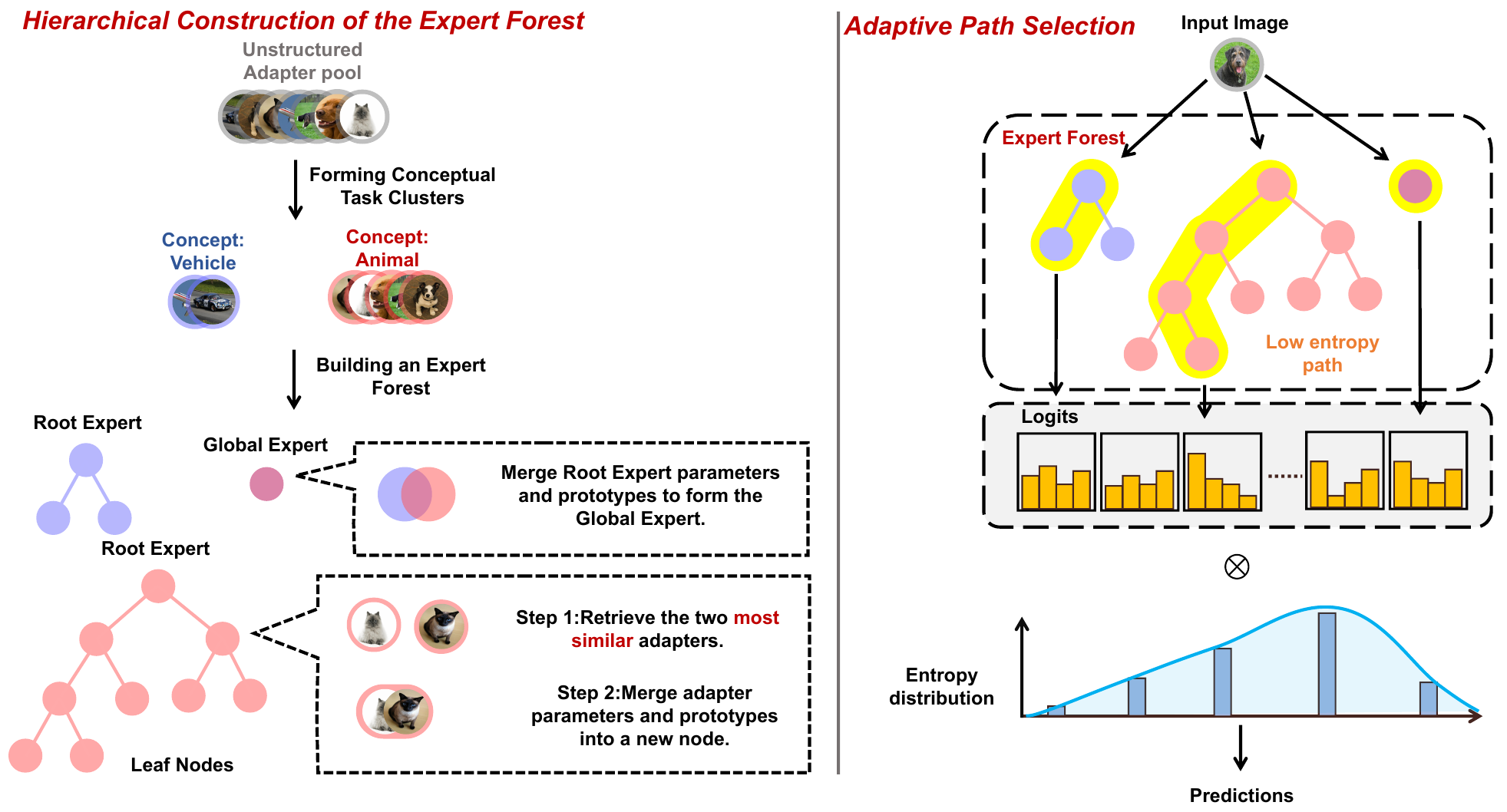}
    \caption{
        Overview of the SAEF. 
        Our method organizes individually trained task adapters into a structured knowledge hierarchy.
        \textbf{(1) Stage 1: Conceptual Clustering.} Based on semantic prototypes, tasks are grouped into high-level conceptual clusters (e.g., 'Animal', 'Vehicle').
        \textbf{(2) Stage 2: Hierarchical Construction.} Within each cluster, a balanced expert tree is built via a bottom-up process of recursively fusing the most visually similar nodes. The roots of these trees are then merged to form a single Global Expert.
        \textbf{(3) Stage 3: Adaptive Inference.} For a given input, SAEF performs a parallel, multi-path search to dynamically activate a set of relevant experts. The final prediction is derived from an entropy-guided fusion of their outputs.
    }
    \label{fig:methodology}
\end{figure*}

\subsection{Our Method: Semantic-guided Adaptive Expert Forest}
\label{sec:our_method}
Our method begins by following the standard PTM-based CIL paradigm, where a separate, task-specific adapter is trained for each incremental task using the objective in Eq.~\eqref{eq:adapter-overall}. However, instead of treating the resulting set of adapters as a simple, flat collection during inference, we propose the Semantic-guided Adaptive Expert Forest (SAEF). SAEF is a method that organizes these isolated adapters into a structured, hierarchical knowledge base as new tasks are learned. This process consists of three main stages, as illustrated in Figure~\ref{fig:methodology}: \textit{Conceptual Clustering}, \textit{Hierarchical Construction}, and \textit{Adaptive Inference}. We describe each stage in detail below.

\paragraph{Conceptual Clustering.}
Recognizing that incremental tasks often fall into distinct conceptual domains (e.g., distinguishing 'animals' from 'vehicles'), the first stage of SAEF partitions the set of all $T$ learned tasks into semantically coherent groups. To achieve this, we first define two complementary prototypes for each task $t$: a semantic prototype $\mathbf{p}_t^{\text{s}}$ and a visual prototype $\mathbf{p}_t^{\text{f}}$. Semantic prototypes are derived using a pre-trained and frozen text encoder, denoted as $g_t$ from CLIP~\cite{radford2021learning}. For each class name $c$ belonging to a task, we construct a corresponding text prompt $t_c$ using the template "a photo of a [CLASS]". The semantic prototype for task $t$ is then the mean of these text embeddings:
\begin{equation}
\label{eq:semantic_prototype}
\mathbf{p}_t^{\text{s}} = \frac{1}{|\mathcal{C}_t|} \sum_{c \in \mathcal{C}_t} g_t(t_c).
\end{equation}
Visual prototypes are extracted using the PTM's feature extractor $\phi$, in conjunction with the adapter from the first incremental task $\mathcal{A}_1$. We use this combination $\phi(\cdot; \mathcal{A}_1)$, because $\mathcal{A}_1$ helps to adapt the general pre-trained features to the specific downstream domain, effectively bridging the domain gap. The visual prototype for task $t$ is the mean feature vector of its training samples:
\begin{equation}
\label{eq:visual_prototype}
\mathbf{p}_t^{\text{f}} = \frac{1}{|\mathcal{D}_t|} \sum_{\mathbf{x}_i \in \mathcal{D}_t} \phi(\mathbf{x}_i; \mathcal{A}_1).
\end{equation}
With the semantic prototypes $\{\mathbf{p}_t^{\text{s}}\}_{t=1}^T$ collected, we apply the K-Means algorithm to partition the set of task indices $\{1, \ldots, T\}$ into $K$ conceptual clusters, $\{G_1, \ldots, G_K\}$. The optimal number of clusters is determined automatically by selecting the value that maximizes the average silhouette score~\cite{shahapure2020cluster} over a predefined search range. This clustering process establishes the 'forest' structure of SAEF. Each cluster $G_k$ provides the initial set of task adapters that will be structured into a distinct 'tree' in the \textit{Hierarchical Construction} stage, and the visual prototypes $\mathbf{p}_t^{\text{f}}$ are used to guide this process.

\paragraph{Hierarchical Construction.}
The initial clustering provides a coarse-grained organization of tasks. To enable more effective, fine-grained knowledge transfer, this second stage constructs a balanced binary tree of experts for each conceptual cluster $G_k$, and then merges their roots into a single global expert. This transforms the initially unstructured collection of adapters into a unified, multi-level hierarchy.

For each cluster $G_k$, we start with a set of leaf experts, where each expert corresponds to a single task $t \in G_k$ and holds its adapter parameters $\theta_t \in \mathbb{R}^D$ and visual prototype $\mathbf{p}_t^{\text{f}}$. The tree is built using a bottom-up, iterative process. At each step, we find the two most similar experts in the current set by computing the cosine similarity between their visual prototypes:
\begin{equation}
\label{eq:similarity_metric}
    \text{sim}(i, j) = \frac{{\mathbf{p}_i^{\text{f}}}^\top \mathbf{p}_j^{\text{f}}}{\|\mathbf{p}_i^{\text{f}}\|_2 \|\mathbf{p}_j^{\text{f}}\|_2},
\end{equation}
where $\|\cdot\|_2$ denotes the L2-norm. Once the most similar pair is identified, a new parent expert is created. The parent's parameter vector, $\theta_p \in \mathbb{R}^D$, is formed by merging the parameters of its two children, $\theta_i$ and $\theta_j$, using a vectorized operation designed to preserve the dominant signal from both:
\begin{equation}
\label{eq:param_merge}
    \theta_p \leftarrow \text{sign}(\theta_i + \theta_j) \odot \max(|\theta_i|, |\theta_j|),
\end{equation}
where $\text{sign}(\cdot)$ is the element-wise sign function, $|\cdot|$ and $\max(\cdot, \cdot)$ are element-wise operations for absolute value and maximum, and $\odot$ denotes the Hadamard product. The visual prototype for the new parent is computed as a weighted average of its children's prototypes, with weights proportional to the number of original leaf tasks each child represents. The two child experts are then replaced by their new parent in the set. This process continues until only a single root expert, $R_k$, remains for the cluster. The detailed procedure for prototype merging and managing the set of experts is provided in Appendix~\ref{app:construction_details}.

Finally, to create a single entry point for inference, we merge the $K$ tree roots into one global root expert $R_G$. The parameters of this global root, $\theta_G \in \mathbb{R}^D$, are formed using a vectorized operation that captures both the consensus and the maximum signal strength from all trees:
\begin{equation}
\label{eq:global_root_merge}
    \theta_G \leftarrow \text{sign}\left(\sum_{k=1}^{K} \theta_{R_k}\right) \odot \max_{k \in \{1, \ldots, K\}} |\theta_{R_k}|,
\end{equation}
where $\text{sign}(\cdot)$ is the element-wise sign function, $\max_{k}$ finds the element-wise maximum across the $K$ root vectors, and $\odot$ denotes the Hadamard product. This completes the \textit{Hierarchical Construction} of the full SAEF hierarchy.

\paragraph{Adaptive Inference.}
SAEF uses a dynamic inference strategy to classify any given input $x$. The strategy finds an optimal path in each of the $K$ trees and then fuses the predictions from all experts along these paths.

For any expert in the hierarchy, represented by its adapter $\mathcal{A}_n$, we first compute its class-wise logits for a given input $x$. Specifically, the logit for each class $c \in \mathcal{Y}_T$ is computed as $s_n(c) = \mathbf{w}_c^\top \phi(x; \mathcal{A}_n)$, where $\mathbf{w}_c$ is the classifier weight for class $c$. These logits are then transformed into a predictive distribution $\mathbf{z}_n$ via the softmax function:
\begin{equation}
\label{eq:expert_prediction}
    z_n(c) = \frac{\exp(s_n(c))}{\sum_{j \in \mathcal{Y}_T} \exp(s_n(j))}.
\end{equation}
The expert's predictive confidence is inversely measured by the entropy of this distribution:
\begin{equation}
\label{eq:shannon_entropy}
    H(z_n) = - \sum_{c \in \mathcal{Y}_T} z_n(c) \log z_n(c).
\end{equation}

The inference process begins with a parallel search, performed independently for each of the $K$ trees. Starting from each tree's root, the search path follows the child expert that shows lower predictive entropy for the given input. This recursive selection defines an optimal inference path $\mathcal{P}_k$ from root to leaf for each tree. Further details on this entropy-guided selection process are provided in Appendix~\ref{app:inference_details}.

The final prediction is derived from an activated set of experts $\mathcal{E}$ which includes the global root $R_G$ and all experts from the $K$ identified paths:
\begin{equation}
\label{eq:activated_set}
    \mathcal{E} = \{R_G\} \cup \bigcup_{k=1}^{K} \mathcal{P}_k.
\end{equation}
The fusion weight $w_n$ for each activated expert is based on its confidence (negative entropy), normalized across the set with a temperature $\tau$:
\begin{equation}
\label{eq:fusion_weights}
w_n = \frac{\exp(-H(z_n) / \tau)}{\sum_{j \in \mathcal{E}} \exp(-H(z_j) / \tau)}.
\end{equation}
The final predictive distribution $z_f$ is the weighted average of the predictions from all activated experts:
\begin{equation}
\label{eq:final_prediction}
    z_{f} = \sum_{n \in \mathcal{E}} w_n \mathbf{z}_n.
\end{equation}
This strategy allows SAEF to use knowledge from different levels of the hierarchy, weighting each expert's opinion by its confidence to make a robust final prediction.
\begin{table*}[t]
\centering
\caption{
    Average ($\bar{\mathcal{A}}$) and final task ($\mathcal{A}_T$) Top-1 accuracy comparison on four challenging class-incremental learning benchmarks. 
    All methods use ViT-B/16-IN21K as the backbone and are evaluated in the exemplar-free setting.
    The best performance in each column is highlighted in \textbf{bold}.
}
\label{tab:main_results}
\begin{tabular*}{\textwidth}{@{\extracolsep{\fill}}l|cc|cc|cc|cc}
    \toprule
    \multicolumn{1}{c|}{\multirow{2}{*}{\textbf{Method}}} & 
    \multicolumn{2}{c|}{\textbf{CIFAR-100}} & 
    \multicolumn{2}{c|}{\textbf{ImageNet-R}} &
    \multicolumn{2}{c|}{\textbf{ImageNet-A}} & 
    \multicolumn{2}{c}{\textbf{ObjectNet}} \\
    \cmidrule(lr){2-3} \cmidrule(lr){4-5} \cmidrule(lr){6-7} \cmidrule(lr){8-9}
    & $\bar{\mathcal{A}}$ (\%) & $\mathcal{A}_T$ (\%) & 
      $\bar{\mathcal{A}}$ (\%) & $\mathcal{A}_T$ (\%) &
      $\bar{\mathcal{A}}$ (\%) & $\mathcal{A}_T$ (\%) &
      $\bar{\mathcal{A}}$ (\%) & $\mathcal{A}_T$ (\%) \\
    \midrule
    Finetune         & 38.90 & 20.17 & 32.32 & 22.78 & 24.28 & 14.51  & 19.14 & 8.73 \\
    \midrule
    L2P~\cite{wang2022learning}            & 85.94 & 79.93 & 75.46 & 69.77 & 49.39 & 41.71 & 63.78 & 52.19 \\
    DualPrompt~\cite{wang2022dualprompt}   & 87.87 & 81.15 & 73.10 & 67.18 & 53.71 & 41.67 & 59.27 & 49.33 \\
    CODA-Prompt~\cite{smith2023coda}     & 89.11 & 81.96 & 77.97 & 72.27 & 53.54 & 42.73 & 66.07 & 53.29 \\
    SLCA~\cite{zhang2023slca}              & 92.49 & 88.55 & 81.17 & 77.00 & 68.66 & 58.74 & 72.55 & 61.30 \\ 
    SSIAT~\cite{tan2024semantically}             & 93.52 & 90.07 & 83.20 & 78.85 & 70.83 & 62.23 & 73.65 & 62.45 \\
    SimpleCIL~\cite{zhou2024revisiting}    & 87.57 & 81.26 & 61.26 & 54.55 & 59.77 & 48.91 & 65.45 & 53.59 \\
    APER + Adapter~\cite{zhou2024revisiting} & 90.65 & 85.15 & 75.82 & 67.95 & 60.47 & 49.37 & 67.18 & 55.24 \\
    RanPAC~\cite{mcdonnell2023ranpac}    & 94.00 & 90.62 & 82.98 & 77.94 & 69.32 & 61.82 & 72.76 & 62.02 \\
    EASE~\cite{zhou2024expandable}         & 91.51 & 85.80 & 81.74 & 76.17 & 65.34 & 55.04 & 70.84 & 57.86 \\
    MOS~\cite{sun2025mos}                  & 93.30 & 89.25 & 82.96 & 77.93 & 67.08 & 56.22 & 74.69 & 63.62 \\
    \midrule
    \textbf{\method (Ours)}                   & \bf{94.53} & \bf{90.60} & \bf{84.54} & \bf{80.02} & \bf{73.39} & \bf{64.91} & \bf{76.38} & \bf{66.58} \\
    \bottomrule
\end{tabular*}
\end{table*}
\begin{figure*}[t!]
\centering

\begin{subfigure}{0.32\textwidth}
    \includegraphics[width=\linewidth]{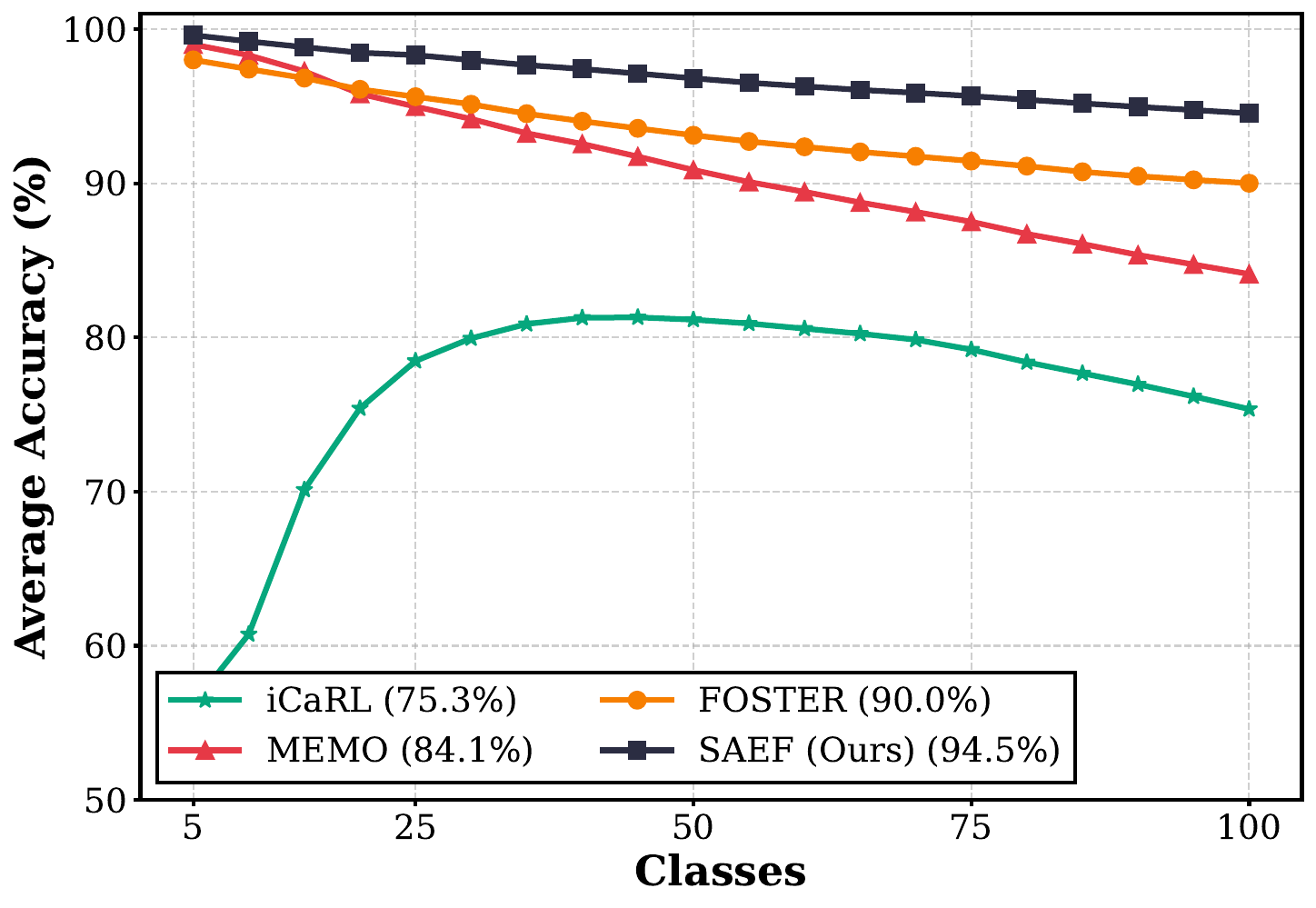} 
    \caption{CIFAR-100 (20 tasks of 5 classes)}
    \label{fig:rehearsal_cifar}
\end{subfigure}
\hfill 
\begin{subfigure}{0.32\textwidth}
    \includegraphics[width=\linewidth]{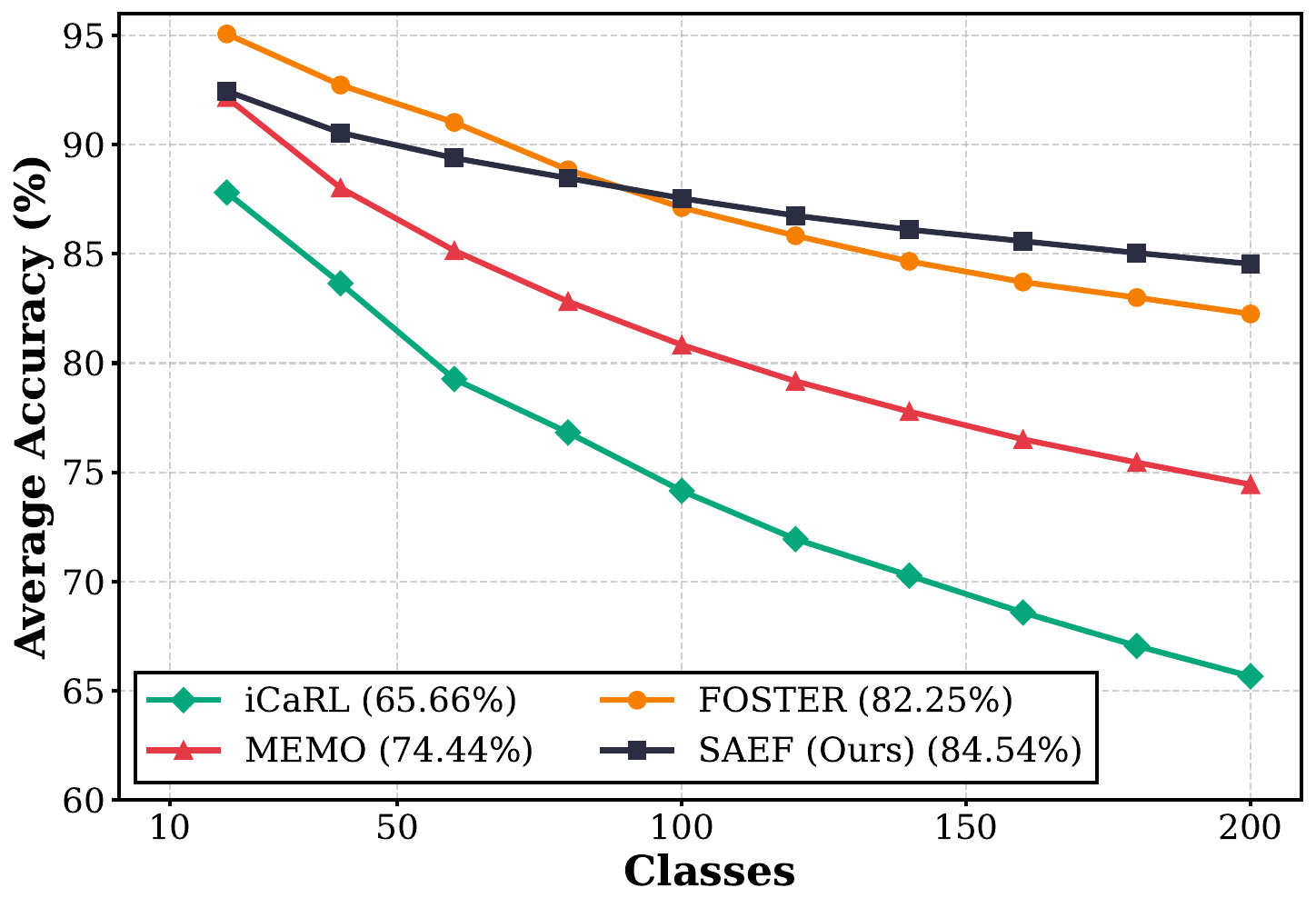} 
    \caption{ImageNet-R (10 tasks of 20 classes)}
    \label{fig:rehearsal_imagenet}
\end{subfigure}
\hfill 
\begin{subfigure}{0.32\textwidth}
    \includegraphics[width=\linewidth]{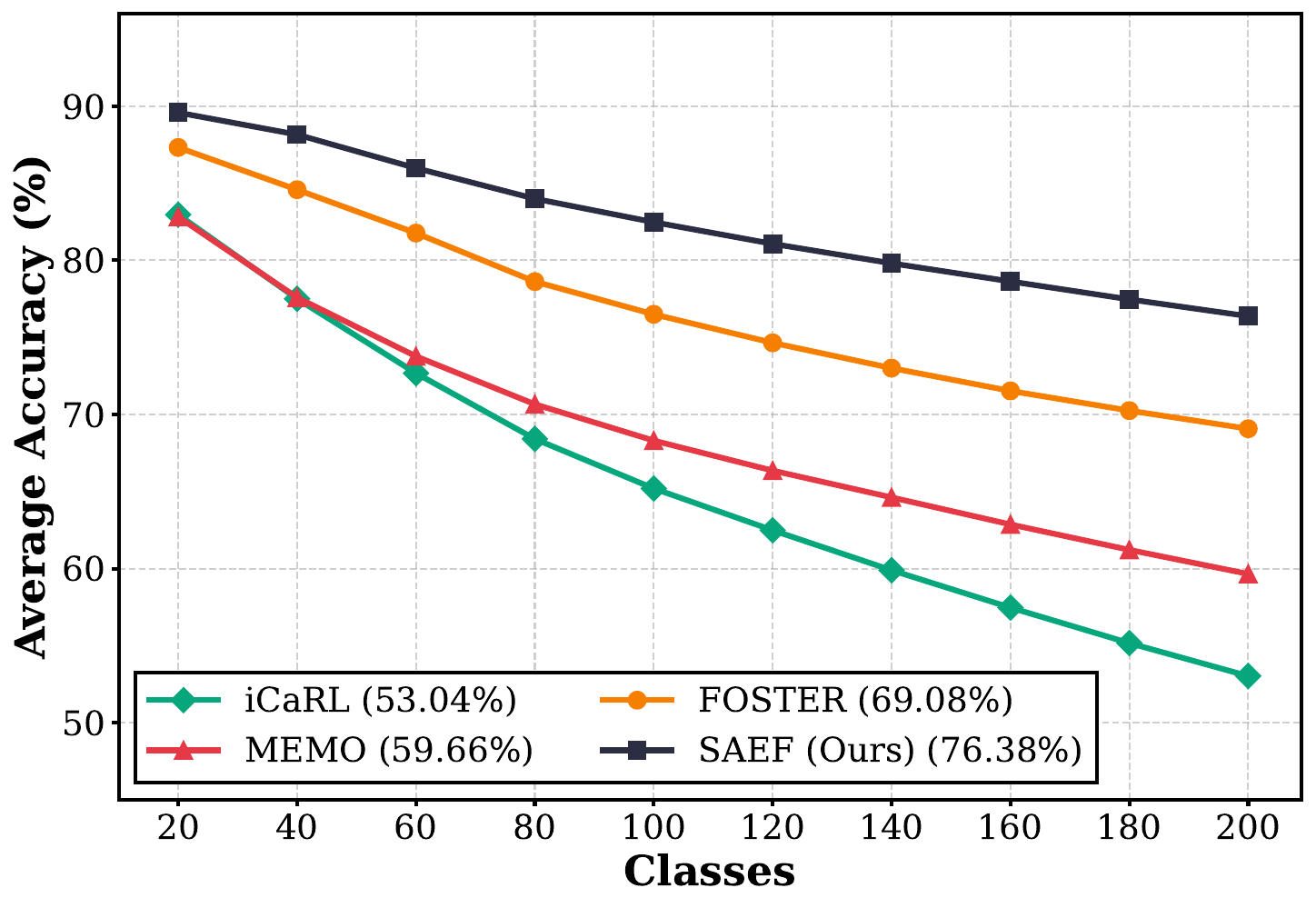} 
    \caption{ObjectNet (10 tasks of 15 classes)}
    \label{fig:rehearsal_objectnet}
\end{subfigure}

\caption{
    Comparison against rehearsal-based methods.
    Average accuracy ($\bar{\mathcal{A}}$) is plotted as a function of the number of learned tasks. Rehearsal-based methods use 20 exemplars per class.
    Despite being an exemplar-free method, SAEF consistently outperforms strong rehearsal-based competitors across multiple benchmarks, and the performance gap often widens over time, demonstrating superior knowledge retention and scalability.
}
\label{fig:rehearsal_comparison}
\end{figure*}
\section{Experiments}
\label{sec:experiments}
\subsection{Implementation Details}

\noindent\textbf{Datasets and Task Protocols.}
To rigorously evaluate SAEF's ability to manage knowledge across diverse domains, we adopt the comprehensive evaluation framework proposed by~\cite{zhou2024revisiting}. Our evaluation is conducted on four key benchmarks: CIFAR-100~\cite{krizhevsky2009learning}, ImageNet-R~\cite{hendrycks2021many}, ImageNet-A~\cite{hendrycks2021natural}, and ObjectNet~\cite{barbu2019objectnet}. This selection deliberately includes not only standard CIL benchmarks but also challenging datasets (ImageNet-A, ObjectNet), which exhibit significant domain shifts relative to the ImageNet pre-training data. For task partitioning, CIFAR-100 (100 classes) is divided into 20 incremental tasks of 5 classes each. The other three datasets, ImageNet-R, ImageNet-A, and ObjectNet, each with 200 classes, are consistently partitioned into 10 tasks of 20 classes each. For a fair and reproducible evaluation, we adopt the protocol from~\cite{rebuffi2017icarl}, where class orders are shuffled using a fixed random seed of 1993 before the data is partitioned into tasks.

\noindent\textbf{Training Details.}
All models are implemented in PyTorch~\cite{paszke2019pytorch}, leveraging the PILOT codebase~\cite{sun2023pilot} for reproducibility. The experiments were conducted on NVIDIA GeForce RTX 2080 Ti GPUs. To ensure a fair comparison, all methods utilize a shared backbone: the ViT-B/16 model pre-trained on ImageNet-21K. 

For SAEF, each incremental task is trained for 20 epochs with a batch size of 48. Optimization is performed using an SGD optimizer with momentum, with an initial learning rate of 0.01 that decays following a cosine annealing schedule. Following the training of all adapters, we configure SAEF's structure and inference. The number of conceptual clusters ($K$) is automatically determined by optimizing the average silhouette score~\cite{shahapure2020cluster} over a range of potential values. The fusion temperature $\tau$ (Eq.~\eqref{eq:fusion_weights}) is set to a default value of 1.0. To maintain consistency across all adapter-based methods, including our own, the intermediate projection dimension for all adapters is fixed at 16.


\noindent\textbf{Comparison Methods.}
To validate the effectiveness of our method, we benchmark SAEF against a comprehensive suite of competing methods. Our primary comparison group consists of prominent PTM-based CIL methods, including L2P~\cite{wang2022learning}, DualPrompt~\cite{wang2022dualprompt}, CODA-Prompt~\cite{smith2023coda}, RanPAC~\cite{mcdonnell2023ranpac}, SimpleCIL~\cite{zhou2024revisiting}, APER~\cite{zhou2024revisiting}, SLCA~\cite{zhang2023slca}, SSIAT~\cite{tan2024semantically}, EASE~\cite{zhou2024expandable}, and MOS~\cite{sun2025mos}. Additionally, we include several influential rehearsal-based methods that have been adapted for the PTM paradigm: FOSTER~\cite{wang2022foster}, MEMO~\cite{zhou2022model}, and iCaRL~\cite{rebuffi2017icarl}. Finally, a 'Finetune' baseline, which sequentially updates the entire PTM on each new task, is included to establish a performance lower bound and quantify the effects of catastrophic forgetting.

\noindent\textbf{Evaluation Metrics.}
We adhere to the standard CIL evaluation protocol from~\cite{rebuffi2017icarl}. Let $A_{i,j}$ denote the accuracy on task $j$ after the model has trained up to task $i$. We report two primary metrics to assess performance. The Average Accuracy ($\bar{\mathcal{A}}$) quantifies the model's performance throughout the entire learning sequence, calculated as the average of mean accuracies after each task is completed: $\bar{\mathcal{A}} = \frac{1}{T} \sum_{t=1}^{T} \left( \frac{1}{t} \sum_{j=1}^{t} A_{t,j} \right)$. The Final Accuracy ($\mathcal{A}_T$) measures the average performance across all tasks once the full training sequence has concluded: $\mathcal{A}_T = \frac{1}{T} \sum_{j=1}^{T} A_{T,j}$. Higher values for both metrics indicate a more effective CIL method.

\subsection{Main Results}

We first evaluate SAEF against SOTA exemplar-free methods, with detailed results in Table~\ref{tab:main_results}. Our method achieves the highest scores in both average accuracy ($\bar{\mathcal{A}}$) and final task accuracy ($\mathcal{A}_T$) across all four benchmarks. SAEF surpasses not only strong prompt-based methods like CODA-Prompt but also SOTA adapter-based competitors. For instance, on the challenging ImageNet-R benchmark, SAEF achieves an average accuracy of 84.54\%, outperforming the strongest competitor SSIAT by a significant 1.34\%. This advantage stems from SAEF's ability to organize experts into a semantic hierarchy, which facilitates targeted positive transfer among conceptually related tasks while preventing negative interference from irrelevant ones. In contrast, methods like MOS and EASE, which treat adapters as an isolated collection, consider all prior knowledge equally relevant. This flat, unstructured method leads to suboptimal and noisy predictions, as evidenced by their performance gaps.

We further compare SAEF with strong rehearsal-based methods to test its scalability, even though SAEF operates in the more difficult exemplar-free setting. We follow~\cite{zhou2024expandable} to set the exemplar number to 20 per class for these methods. As shown in Figure~\ref{fig:rehearsal_comparison}, SAEF still outperforms these powerful competitors. On the long 10-task ObjectNet sequence, SAEF's final average accuracy reaches 76.38\%, surpassing the strong FOSTER baseline by a substantial 7.3\%. Critically, the performance gap between SAEF and rehearsal-based methods widens as more tasks are learned. This result strongly suggests that intelligently structuring and retrieving knowledge is a more scalable strategy for long-sequence CIL than relying on instance replay.

\begin{figure}[t!]
\centering

\begin{subfigure}[b]{0.48\linewidth}
    \includegraphics[width=\linewidth]{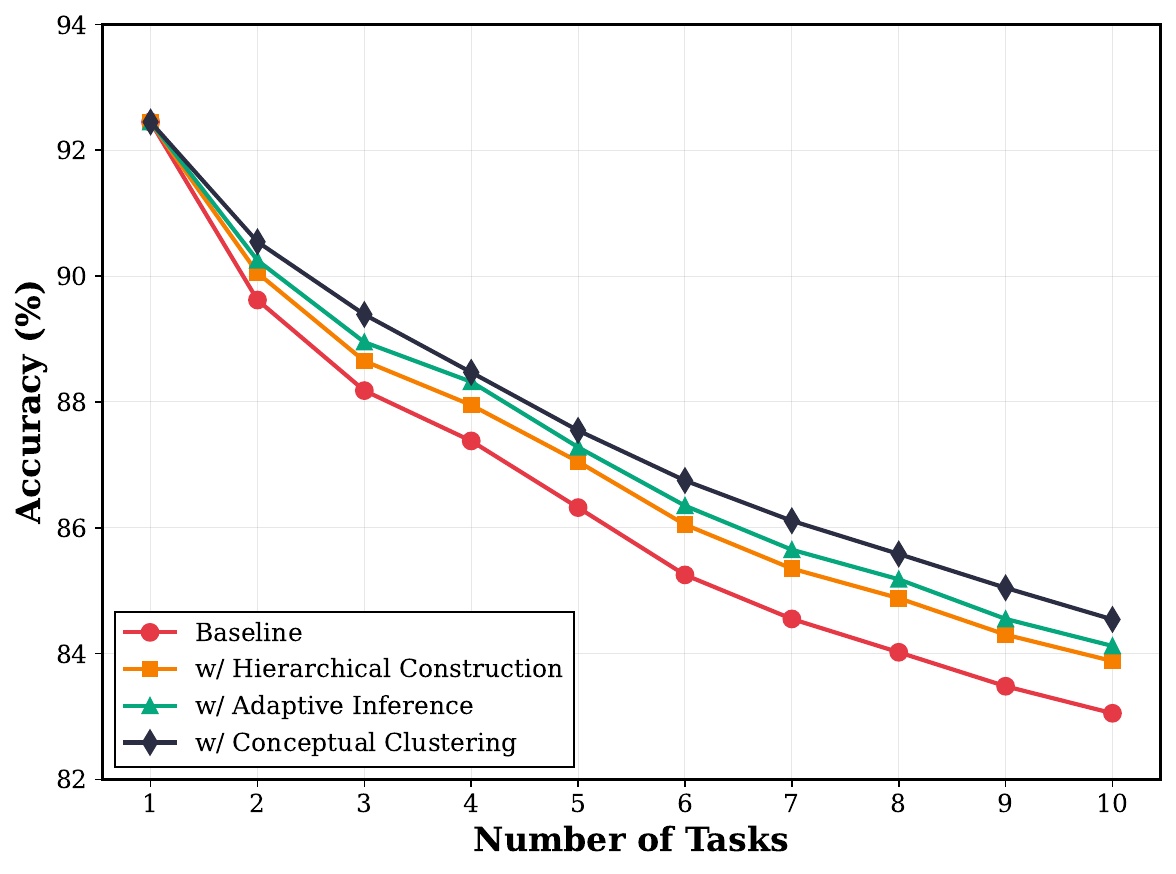} 
    \caption{Ablation of SAEF Components}
    \label{fig:ablation_sub}
\end{subfigure}
\hfill 
\begin{subfigure}[b]{0.48\linewidth}
    \includegraphics[width=\linewidth]{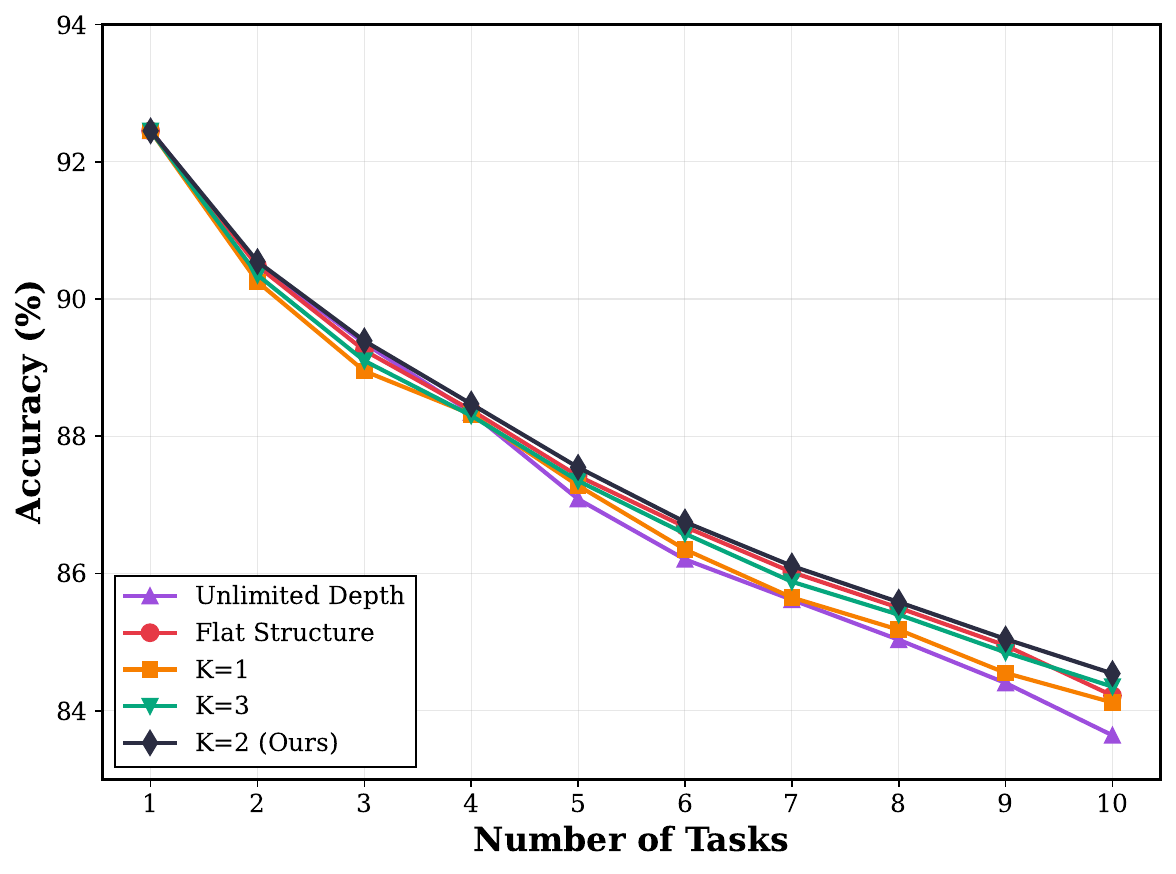} 
    \caption{Analysis of Forest Structure}
    \label{fig:structure_sub}
\end{subfigure}

\caption{
    Ablation and structural analysis of SAEF on ImageNet-R.
    Performance is measured by average accuracy ($\bar{\mathcal{A}}$).
    \textbf{(a)} Ablation study demonstrating the progressive contribution of each core component.
    \textbf{(b)} Structural analysis validating our balanced forest structure ($K=2$) over less structured alternatives.
}
\label{fig:ablation_and_structure}
\end{figure}
\subsection{Ablation Study}

To dissect the contribution of each key component in SAEF, we conduct a progressive ablation study on ImageNet-R, with results visualized in Figure~\ref{fig:ablation_and_structure}(a). Our analysis begins with a 'Baseline' that trains task-specific adapters and selects the final prediction via the maximum logit from a full ensemble. Building upon this, we incrementally propose our components.
First, organizing the flat adapter collection into a single hierarchical tree ('w/ Hierarchical Construction') yields an immediate performance uplift. This demonstrates the inherent value of structured knowledge representation, as it allows for the creation of generalized experts at internal nodes, even without semantic pre-sorting. Next, incorporating our entropy-driven adaptive search and fusion strategy ('w/ Adaptive Inference') delivers a more substantial gain. This highlights the advantage of dynamically activating only the most relevant experts for a given sample, effectively pruning noisy or irrelevant knowledge paths.
Finally, the full SAEF model is formed by introducing the initial stage of semantic clustering ('w/ Conceptual Clustering'), which achieves the highest performance. This final improvement reveals a crucial synergy: pre-grouping tasks into broad conceptual domains (e.g., 'animals', 'vehicles') prevents semantically disparate experts from being forced into the same local hierarchy. This initial partitioning reduces noise and enhances the precision of the subsequent fine-grained knowledge fusion within each tree. Each component is thus shown to play a distinct and synergistic role, culminating in the robust performance of the full SAEF method.

\subsection{Discussion}

\paragraph{Impact of the Expert Forest's Structure.}
We analyze the impact of key structural design choices on SAEF's performance, with results on ImageNet-R presented in Figure~\ref{fig:ablation_and_structure}(b). Our analysis first examines the number of conceptual clusters, $K$. The results show that our default configuration ($K=2$), automatically determined via the silhouette coefficient, achieves the optimal performance. Deviating from this, both a single monolithic tree ($K=1$) and a completely 'Flat Structure' ($K=T$, where each task is its own cluster) impair performance. This suggests that knowledge can be organized in ways that are either too coarse or too fragmented, empirically validating our automated strategy for identifying the ideal number of conceptual domains.

We also evaluate an alternative tree construction strategy: 'Unlimited Depth', which greedily merges the two most similar nodes regardless of their position in the hierarchy. This method yields the poorest results, confirming that merging concepts from disparate hierarchical levels introduces significant semantic noise and disrupts meaningful knowledge abstraction. In contrast, our method's systematic, bottom-up construction preserves the semantic hierarchy by prioritizing merges between nodes at the same hierarchical level. From an efficiency standpoint, both the 'Flat Structure' and 'Unlimited Depth' strategies necessitate a search over all $N$ experts, resulting in $O(N)$ inference complexity. Our balanced forest structure, however, enables a much faster logarithmic search within each tree ($O(\log N)$). This dual superiority in both accuracy and efficiency underscores the critical role of SAEF's carefully designed hierarchical organization.
\begin{table}[t!]
\centering
\setlength{\tabcolsep}{4pt} 
\caption{
    Robustness analysis of the fusion temperature $\tau$ on ImageNet-R. The optimal results are marked in bold.
}
\label{tab:robustness_tau}
\begin{tabular}{@{}lcccc@{}}
\toprule
\textbf{Temperature ($\tau$)} & 1.0  & 0.5 & 0.2 & 0.1 \\ \midrule
\textbf{$\bar{\mathcal{A}}$ (\%)} & \textbf{84.54} & 84.49 & 84.32 & 84.31 \\
\textbf{$\mathcal{A}_T$ (\%)} & 80.02 & \textbf{80.08} & 79.78 & 79.87 \\ \bottomrule
\end{tabular}
\end{table}
\begin{table}[t!]
\centering
\caption{
    Analysis of the trade-off between inference efficiency and average accuracy by varying the entropy threshold $\tau_e$ on CIFAR-100. 
    The 'Flat Structure' baseline represents a non-hierarchical method where all task adapters are queried. 
    The setting of $\tau_e=1.0$ (highlighted in \textbf{bold}) strikes an excellent balance.
}
\label{tab:ablation_entropy_tradeoff}
\resizebox{\columnwidth}{!}{%
\begin{tabular}{lcccc}
    \toprule
    \textbf{Configuration} & \textbf{$\bar{\mathcal{A}}$ (\%) $\uparrow$} & \textbf{Avg. Depth $\downarrow$} & \textbf{Theo. Speedup $\uparrow$} & \textbf{Time (ms) $\downarrow$} \\
    \midrule
    Flat Structure (Baseline) & 94.44 & -- & 1.00$\times$ & 2775.43 \\
    \midrule
    SAEF ($\tau_e = 0.0$) & 94.53 & 4.04 & 2.20$\times$ & 1346.51 \\
    \textbf{SAEF ($\tau_e = 1.0$)} & \textbf{94.24} & \textbf{1.19} & \textbf{5.92$\times$} & \textbf{500.39} \\
    SAEF ($\tau_e = 2.0$) & 93.96 & 1.00 & 6.67$\times$ & 443.00 \\
    \bottomrule
\end{tabular}%
}
\end{table}
\paragraph{Parameter Robustness.}
SAEF's fusion mechanism involves a single key hyperparameter: the temperature $\tau$ in the entropy-weighted softmax (Eq.~\eqref{eq:fusion_weights}), which controls the sharpness of the weighting distribution. To evaluate its robustness, we conducted an analysis on the challenging ImageNet-R benchmark by varying $\tau$ across a range of values: $\{1.0, 0.5, 0.2, 0.1\}$. The results, presented in Table~\ref{tab:robustness_tau}, demonstrate that SAEF's performance is remarkably stable.

\paragraph{Trade-off between Inference Efficiency and Accuracy.}
A key advantage of SAEF's hierarchical structure is the ability to control the trade-off between computational cost and predictive accuracy during inference. This is achieved via an entropy threshold, $\tau_e$, which enables an early-exit mechanism: the search down a path terminates if the current node's predictive entropy falls below $\tau_e$. A higher threshold encourages more frequent early exits, thereby reducing the average search depth and accelerating inference. As shown in Table~\ref{tab:ablation_entropy_tradeoff}, this mechanism allows for significant efficiency gains. Compared to the 'Flat Structure' baseline, which must query all $N$ task adapters, SAEF offers a flexible spectrum of operating points. Setting $\tau_e=1.0$ strikes an excellent balance: it reduces the average search depth to just 1.19 and achieves a nearly 6$\times$ theoretical speedup, while incurring only a negligible 0.29\% drop in accuracy. Notably, the measured inference time reduction closely aligns with our theoretical speedup calculations (see Appendix~\ref{app:speedup_calc}), confirming the practical benefits of our method. This demonstrates that for many samples, high-level, generalized experts in the upper tiers of the forest are sufficient for confident prediction, making SAEF not only effective but also highly practical for real-world deployment.

\begin{figure}[t!]
    \centering
    \includegraphics[width=0.9\columnwidth]{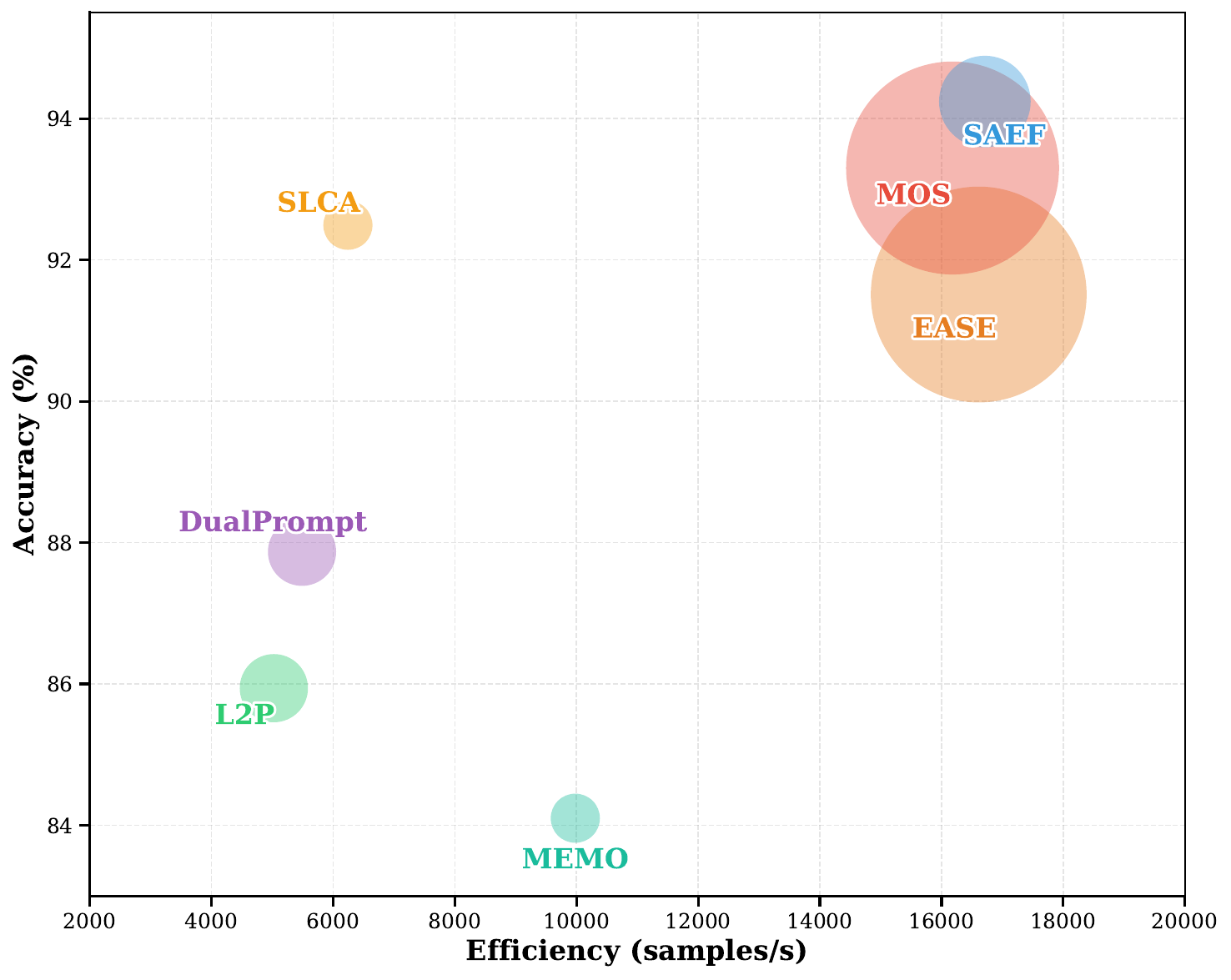} 
    \caption{
        efficiency and performance evaluation on CIFAR-100.
        We plot average accuracy ($\bar{\mathcal{A}}$) against Training Throughput (samples/s). 
        The size of each bubble is proportional to the Inference Time (ms/sample). 
        The ideal method occupies the top-right corner with the smallest bubble, representing high accuracy, fast training, and rapid inference.
    }
    \label{fig:efficiency_tradeoff}
\end{figure}

\paragraph{Efficiency Evaluation.}
To provide a holistic efficiency and performance comparison, we present a multi-faceted analysis in Figure~\ref{fig:efficiency_tradeoff}, benchmarked on CIFAR-100. Prompt-based methods like L2P and DualPrompt occupy the lower-left region, characterized by moderate training throughput and relatively fast inference (indicated by their smaller bubbles), albeit with lower accuracy. In contrast, adapter-based methods generally offer a compelling combination of high accuracy and efficient training throughput. However, a critical distinction emerges at inference time. Methods like EASE and MOS are hindered by severe inference latency, a direct consequence of their need to query most adapters in the pool to gather logits. This is reflected in their large bubble sizes, rendering them impractical for real-time applications.

SAEF ($\tau_e=1.0$) carves out a unique and highly desirable position in this landscape. It is firmly located in the coveted top-right corner, achieving the highest accuracy (94.24\%) among all competitors while maintaining a high training throughput comparable to other leading adapter methods. Crucially, its adaptive inference strategy keeps its bubble size remarkably small, demonstrating an inference time that is over 5.5$\times$ faster than full-ensemble methods like EASE. This unique combination of SOTA accuracy, high training throughput, and rapid inference underscores SAEF's superiority as a practical and scalable solution for real-world CIL scenarios.
\section{Conclusion}
\label{sec:conclusion}

\noindent
Effectively reusing knowledge in adapter-based CIL remains a significant challenge.
While adapters prevent catastrophic forgetting, their treatment as an unstructured, flat collection leads to suboptimal, indiscriminate knowledge sharing.
To address this, we proposed the SAEF, a method that shifts the focus from treating adapters as an isolated set to organizing them into a semantic hierarchy.
SAEF operates by automatically constructing a forest of balanced expert trees, where tasks are grouped by semantic relatedness and progressively merged.
This structure allows the model to perform targeted knowledge transfer by activating only relevant expert paths, preventing interference from conceptually distant tasks.
Extensive experiments on major CIL benchmarks demonstrate that SAEF achieves superior accuracy over current SOTA methods, validating the effectiveness of hierarchical knowledge organization for CIL.
{
    \small
    \bibliographystyle{ieeenat_fullname}
    \bibliography{main}
}
\clearpage
\setcounter{page}{1}
\maketitlesupplementary
\appendix
\setcounter{table}{0}
\renewcommand{\thetable}{A\arabic{table}}
\setcounter{figure}{0}
\renewcommand{\thefigure}{A\arabic{figure}}

\section{Detailed Implementation of SAEF Operations}
\label{app:details}
In this appendix, we provide the detailed implementation specifics for the \textit{Hierarchical Construction} and \textit{Adaptive Inference} stages, which are described at a higher level of abstraction in the main text. We also detail the prototype-based classifier and its alignment mechanism, which are foundational to our method.
\begin{algorithm*}[t!]
\caption{SAEF: Offline Hierarchy Construction}
\label{alg:saef_construction}
\begin{algorithmic}[1]
\Statex \textbf{Inputs:}
\Statex \quad Set of all learned task adapters $\{\mathcal{A}_t\}_{t=1}^T$ with parameters $\{\theta_t\}_{t=1}^T$
\Statex \quad Training data for all tasks $\{\mathcal{D}_t\}_{t=1}^T$
\Statex \quad Frozen text encoder $g_t$ (from CLIP), frozen feature extractor $\phi$ (from ViT)
\Statex \quad Search range for number of clusters $K_{range}$
\Statex \textbf{Output:} The complete expert hierarchy, `Hierarchy`$=(\mathcal{R}, R_G)$

\vspace{0.2cm}
\Procedure{BuildHierarchy}{$\{\mathcal{A}_t\}, \{\mathcal{D}_t\}, g_t, \phi, K_{range}$}
    \Statex \Comment{--- \textbf{Stage 1: Conceptual Clustering} ---}
    \State Initialize prototype sets: $\mathcal{P}_s \leftarrow \emptyset$, $\mathcal{P}_f \leftarrow \emptyset$
    \For{$t = 1 \to T$} \Comment{Compute prototypes for all tasks}
        \State $\mathbf{p}_t^{\text{s}} \leftarrow \frac{1}{|\mathcal{C}_t|} \sum_{c \in \mathcal{C}_t} g_t(\text{"a photo of a [c]"})$ \Comment{Eq.~\eqref{eq:semantic_prototype}}
        \State $\mathbf{p}_t^{\text{f}} \leftarrow \frac{1}{|\mathcal{D}_t|} \sum_{\mathbf{x}_i \in \mathcal{D}_t} \phi(\mathbf{x}_i; \mathcal{A}_1)$ \Comment{Eq.~\eqref{eq:visual_prototype}}
        \State Add $\mathbf{p}_t^{\text{s}}$ to $\mathcal{P}_s$; Add $\mathbf{p}_t^{\text{f}}$ to $\mathcal{P}_f$
    \EndFor
    \State $K^* \leftarrow \text{FindOptimalK}(\mathcal{P}_s, K_{range})$ \Comment{Using silhouette score}
    \State $\{G_1, \ldots, G_{K^*}\} \leftarrow \text{KMeans}(\mathcal{P}_s, K^*)$ \Comment{Partition tasks into $K^*$ clusters}

    \Statex
    \Statex \Comment{--- \textbf{Stage 2: Hierarchical Construction} ---}
    \State Initialize tree root set: $\mathcal{R} \leftarrow \emptyset$
    \For{$k = 1 \to K^*$}
        \State $R_k \leftarrow \text{BuildTree}(G_k, \{\mathcal{A}_t\}_{t \in G_k}, \{\mathbf{p}_t^{\text{f}}\}_{t \in G_k})$
        \State Add $R_k$ to $\mathcal{R}$
    \EndFor
    \State $\theta_G \leftarrow \text{sign}(\sum_{R_k \in \mathcal{R}} \theta_{R_k}) \odot \max_{R_k \in \mathcal{R}} |\theta_{R_k}|$ \Comment{Merge tree roots to form Global Expert, Eq.~\eqref{eq:global_root_merge}}
    \State $R_G \leftarrow \text{NewNode}(\mathcal{A}_G(\theta_G))$
    \State \textbf{return} Hierarchy: $(\mathcal{R}, R_G)$
\EndProcedure

\vspace{0.3cm}
\Procedure{BuildTree}{Cluster $G_k$, Adapters $\{\mathcal{A}_t\}$, Prototypes $\{\mathbf{p}_t^{\text{f}}\}$}
    \State Initialize working set $\mathcal{N}$ with leaf experts from cluster $G_k$. Each node $n_t$ contains $(\mathcal{A}_t, \mathbf{p}_t^{\text{f}})$.
    \While{$|\mathcal{N}| > 1$}
        \State $(n_i, n_j) \leftarrow \mathop{\arg\max}_{n_a, n_b \in \mathcal{N}} \text{sim}(\mathbf{p}_a^{\text{f}}, \mathbf{p}_b^{\text{f}})$ \Comment{Find most similar pair using Eq.~\eqref{eq:similarity_metric}}
        \State $\theta_p \leftarrow \text{sign}(\theta_i + \theta_j) \odot \max(|\theta_i|, |\theta_j|)$ \Comment{Merge parameters using Eq.~\eqref{eq:param_merge}}
        \State $\mathbf{p}_p^{\text{f}} \leftarrow \text{WeightedAvgMerge}(\mathbf{p}_i^{\text{f}}, \mathbf{p}_j^{\text{f}})$ \Comment{Merge prototypes, see Appx.~\ref{app:construction_details}}
        \State $n_p \leftarrow \text{NewNode}(\mathcal{A}_p(\theta_p), \mathbf{p}_p^{\text{f}}, \text{children}=(n_i, n_j))$
        \State Remove $n_i, n_j$ from $\mathcal{N}$; Add $n_p$ to $\mathcal{N}$
    \EndWhile
    \State \textbf{return} The single remaining node in $\mathcal{N}$ (the tree root)
\EndProcedure
\end{algorithmic}
\end{algorithm*}

\begin{algorithm*}[t!]
\caption{SAEF: Online Adaptive Inference}
\label{alg:saef_inference}
\begin{algorithmic}[1]
\Statex \textbf{Inputs:}
\Statex \quad Test sample $x$
\Statex \quad The pre-built `Hierarchy`$=(\mathcal{R}, R_G)$ from Algorithm~\ref{alg:saef_construction}
\Statex \quad Frozen feature extractor $\phi$, Classifier $W_T$
\Statex \quad Fusion temperature $\tau$, Entropy threshold $\tau_e$
\Statex \textbf{Output:} The final predicted class label $y_{pred}$

\vspace{0.2cm}
\Procedure{Inference}{$x, (\mathcal{R}, R_G), \phi, W_T, \tau, \tau_e$}
    \Statex \Comment{--- \textbf{Stage 3: Adaptive Inference} ---}
    \State Initialize activated expert set: $\mathcal{E} \leftarrow \{R_G\}$ \Comment{Start with the Global Expert}
    \For{each tree root $R_k \in \mathcal{R}$}
        \State $\mathcal{P}_k \leftarrow \text{FindPath}(R_k, x, \phi, W_T, \tau_e)$ \Comment{Find low-entropy path in each tree}
        \State $\mathcal{E} \leftarrow \mathcal{E} \cup \mathcal{P}_k$ \Comment{Collect all experts on the paths, Eq.~\eqref{eq:activated_set}}
    \EndFor
    
    \State Initialize results list: $\mathcal{S} \leftarrow []$
    \For{each expert $n \in \mathcal{E}$}
        \State $s_n \leftarrow W_T^\top \phi(x; \mathcal{A}_n)$ \Comment{Compute logits for all classes $\mathcal{Y}_T$}
        \State $\mathbf{z}_n \leftarrow \text{Softmax}(s_n)$ \Comment{Get prediction distribution, Eq.~\eqref{eq:expert_prediction}}
        \State $H_n \leftarrow -\sum_c z_n(c) \log z_n(c)$ \Comment{Calculate confidence via entropy, Eq.~\eqref{eq:shannon_entropy}}
        \State Add $(\mathbf{z}_n, H_n)$ to $\mathcal{S}$
    \EndFor
    
    \State $Z_{sum} \leftarrow \sum_{(\mathbf{z}_j, H_j) \in \mathcal{S}} \exp(-H_j / \tau)$ \Comment{Compute normalization term for weights}
    \State $\mathbf{z}_{f} \leftarrow \mathbf{0}$
    \For{each $(\mathbf{z}_n, H_n) \in \mathcal{S}$}
        \State $w_n \leftarrow \exp(-H_n / \tau) / Z_{sum}$ \Comment{Compute fusion weight, Eq.~\eqref{eq:fusion_weights}}
        \State $\mathbf{z}_{f} \leftarrow \mathbf{z}_{f} + w_n \cdot \mathbf{z}_n$ \Comment{Aggregate predictions, Eq.~\eqref{eq:final_prediction}}
    \EndFor
    \State \textbf{return} $\mathop{\arg\max} \mathbf{z}_{f}$
\EndProcedure

\vspace{0.3cm}
\Procedure{FindPath}{$node, x, \phi, W_T, \tau_e$} \Comment{Recursive helper function for path selection}
    \State $s_{node} \leftarrow W_T^\top \phi(x; \mathcal{A}_{node})$
    \State $H_{node} \leftarrow \text{Entropy}(\text{Softmax}(s_{node}))$
    \If{$node$ is a leaf} \Comment{Early exit if confident or at a leaf}
        \State \textbf{return} $\{node\}$
    \Else
        \State $(n_L, n_R) \leftarrow \text{children of } node$
        \State $s_L \leftarrow W_T^\top \phi(x; \mathcal{A}_L)$; \quad $H_L \leftarrow \text{Entropy}(\text{Softmax}(s_L))$
        \State $s_R \leftarrow W_T^\top \phi(x; \mathcal{A}_R)$; \quad $H_R \leftarrow \text{Entropy}(\text{Softmax}(s_R))$
        \If{$H_L < H_R$}
            \State \textbf{return} $\{node\} \cup \text{FindPath}(n_L, x, \phi, W_T, \tau_e)$
        \Else
            \State \textbf{return} $\{node\} \cup \text{FindPath}(n_R, x, \phi, W_T, \tau_e)$
        \EndIf
    \EndIf
\EndProcedure
\end{algorithmic}
\end{algorithm*}
\subsection{Hierarchical Construction Details}
\label{app:construction_details}

The tree construction process operates on a set of nodes, where each expert in our hierarchy is represented as a node. Each node is a data structure containing its expert adapter, its visual prototype, and pointers to its children (if any).

\paragraph{Node Representation.} A leaf node, corresponding to an initial task $t$, is initialized as $node_t = (\mathcal{A}_t, \mathbf{p}_t^{\text{f}})$. A parent node $node_p$, created by merging two children ($node_i$ and $node_j$), has the structure $node_p = (\mathcal{A}_p, \mathbf{p}_p^{\text{f}}, node_i, node_j)$.

\paragraph{Node Selection.} As mentioned in the main text, tree construction is an iterative process that operates on a working set of nodes, $\mathcal{N}_k$, for each cluster $G_k$. At each step, we identify the two most visually similar nodes, $node_i^*$ and $node_j^*$, by selecting the pair from $\mathcal{N}_k$ that maximizes the cosine similarity between their visual prototypes. The formal selection rule is:
\begin{equation}
\label{eq:node_selection_appendix}
    (node_i^*, node_j^*) = \mathop{\arg\max}_{node_i, node_j \in \mathcal{N}_k, i \neq j} \frac{{\mathbf{p}_i^{\text{f}}}^\top \mathbf{p}_j^{\text{f}}}{\|\mathbf{p}_i^{\text{f}}\|_2 \|\mathbf{p}_j^{\text{f}}\|_2}.
\end{equation}

\paragraph{Prototype Merging.} When a new parent node is created from children $node_i^*$ and $node_j^*$, its visual prototype $\mathbf{p}_p^{\text{f}}$ is computed as a weighted average of its children's prototypes. The weights are proportional to the number of original leaf-level tasks that each child node represents. Let $C(node)$ be the set of all leaf tasks that are descendants of a given $node$. The parent prototype is then formally defined as:
\begin{equation}
\label{eq:prototype_merge_appendix}
    \mathbf{p}_p^{\text{f}} = \frac{|C(node_i^*)| \cdot \mathbf{p}_{i^*}^{\text{f}} + |C(node_j^*)| \cdot \mathbf{p}_{j^*}^{\text{f}}}{|C(node_i^*)| + |C(node_j^*)|}.
\end{equation}
This weighting scheme ensures that the parent's prototype is more influenced by the child that represents a larger body of knowledge. After the merge, $node_i^*$ and $node_j^*$ are removed from the working set $\mathcal{N}_k$, and the new parent node is added.

\subsection{Adaptive Inference Details}
\label{app:inference_details}

\paragraph{Entropy-Guided Path Selection.} During inference, SAEF performs a search to find the optimal path in each of the $K$ trees. The search begins at the root of a tree and recursively descends to one of its children. At any non-leaf node (indexed by $p$) with children (indexed by $L$ and $R$ for left and right), the search proceeds to the child node that exhibits lower predictive entropy for the given input sample $x$. The formal rule for selecting the next node in the path is:
\begin{equation}
\label{eq:node_search_appendix}
    node_{\text{next}} = \mathop{\arg\min}_{n \in \{node_L, node_R\}} H(\mathbf{z}_n),
\end{equation}
where $H(\mathbf{z}_n)$ is the Shannon entropy of the prediction from the expert at node $n$, as defined in Eq.~\eqref{eq:shannon_entropy}. This process is repeated until a leaf node is reached, defining a complete path from root to leaf.

\subsection{Prototype-based Classifier and Alignment}
\label{app:alignment}

Our method relies on a non-parametric classifier constructed from class prototypes. Here, we detail how these prototypes are created and aligned across different task-specific feature spaces.

\paragraph{Classifier Construction for New Classes.} For any new class $c$ introduced in the current task $t$ with training samples $\mathcal{D}_c$, its prototype $\mathbf{p}_c$ is computed as the mean feature vector of its samples, extracted using the current task's adapter $\mathcal{A}_t$:
\begin{equation}
\label{eq:app_prototype_calc}
    \mathbf{p}_c = \frac{1}{|\mathcal{D}_c|} \sum_{x \in \mathcal{D}_c} \phi(x; \mathcal{A}_t).
\end{equation}
The corresponding classifier weight $\mathbf{w}_c$ is simply its L2-normalized prototype: $\mathbf{w}_c = \mathbf{p}_c / \|\mathbf{p}_c\|_2$.

\paragraph{Classifier Alignment for Old Classes.} This non-parametric method, however, introduces the subspace misalignment problem: prototypes computed for different tasks reside in distinct, incompatible feature subspaces generated by their respective adapters. To overcome this challenge without storing past exemplars, we adopt a widely-used alignment strategy inspired by methods like MOS~\cite{sun2025mos}. The process involves two steps:

\begin{enumerate}
    \item \textbf{Store Distributional Statistics:} For each past class $c'$, we model and store its feature distribution as a multivariate Gaussian, $\mathcal{N}(\mu_{c'}, \Sigma_{c'})$. These statistics are computed from the features extracted during that class's initial training.
    
    \item \textbf{Re-estimate Prototypes in New Subspace:} When learning a new task $t$, we generate pseudo-features by sampling from these stored distributions. These pseudo-features are then projected through the new task's adapter, $\mathcal{A}_t$, to re-estimate the prototypes for all past classes within the new, unified subspace.
\end{enumerate}

This procedure yields a fully aligned classifier head, $W_t = [\mathbf{w}_1, \ldots, \mathbf{w}_{|\mathcal{Y}_t|}]$, where all class weights are comparable within the same feature space, enabling a fair comparison of logits during prediction.

\section{SAEF Algorithm Pseudocode}
\label{app:pseudocode}

To provide a clear and procedural overview of our method, we present the SAEF method as two separate algorithms. Algorithm~\ref{alg:saef_construction} details the offline \textbf{hierarchy construction} phase, which is performed once after all tasks have been learned. Algorithm~\ref{alg:saef_inference} details the online \textbf{adaptive inference} phase, which is executed for each test sample.

\begin{figure*}[t!]
    \centering
    \begin{subfigure}[b]{0.48\textwidth}
        \centering
        \includegraphics[width=\textwidth]{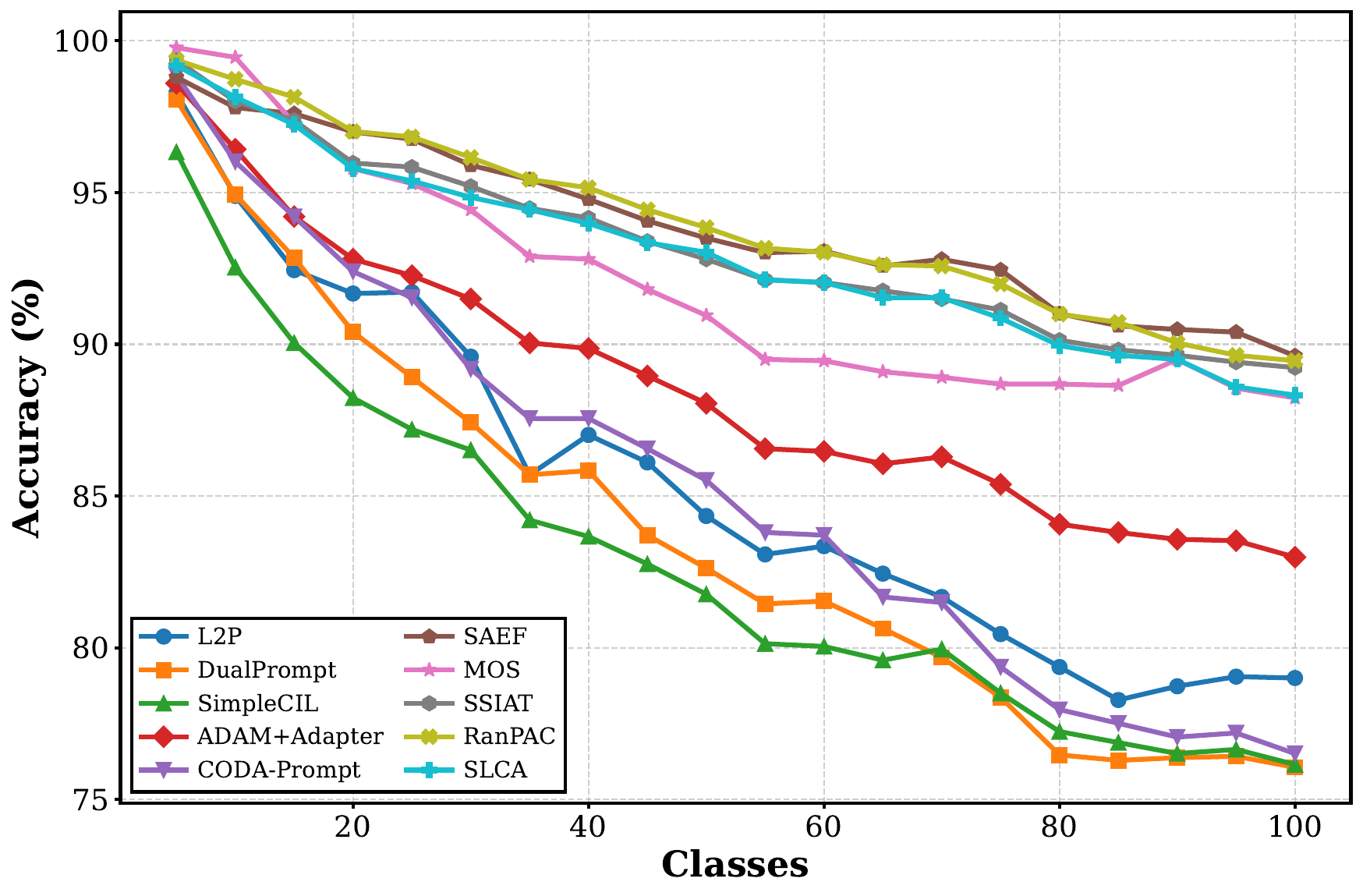}
        \caption{CIFAR-100}
        \label{subfig:appendix_curve_cifar100}
    \end{subfigure}
    \hfill 
    \begin{subfigure}[b]{0.48\textwidth}
        \centering
        \includegraphics[width=\textwidth]{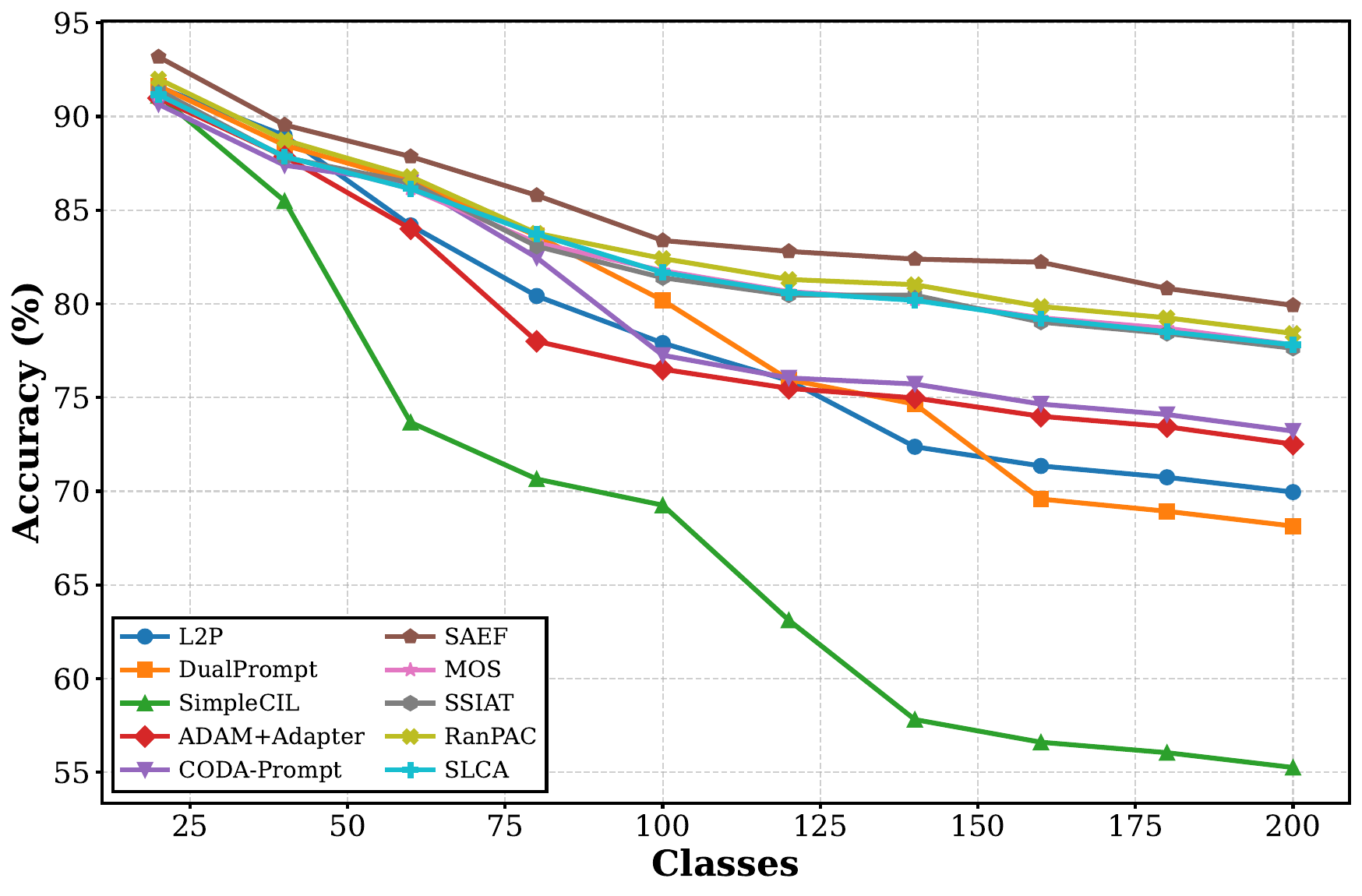}
        \caption{ImageNet-R}
        \label{subfig:appendix_curve_imagenetr}
    \end{subfigure}

    \vspace{0.5cm} 

    \begin{subfigure}[b]{0.48\textwidth}
        \centering
        \includegraphics[width=\textwidth]{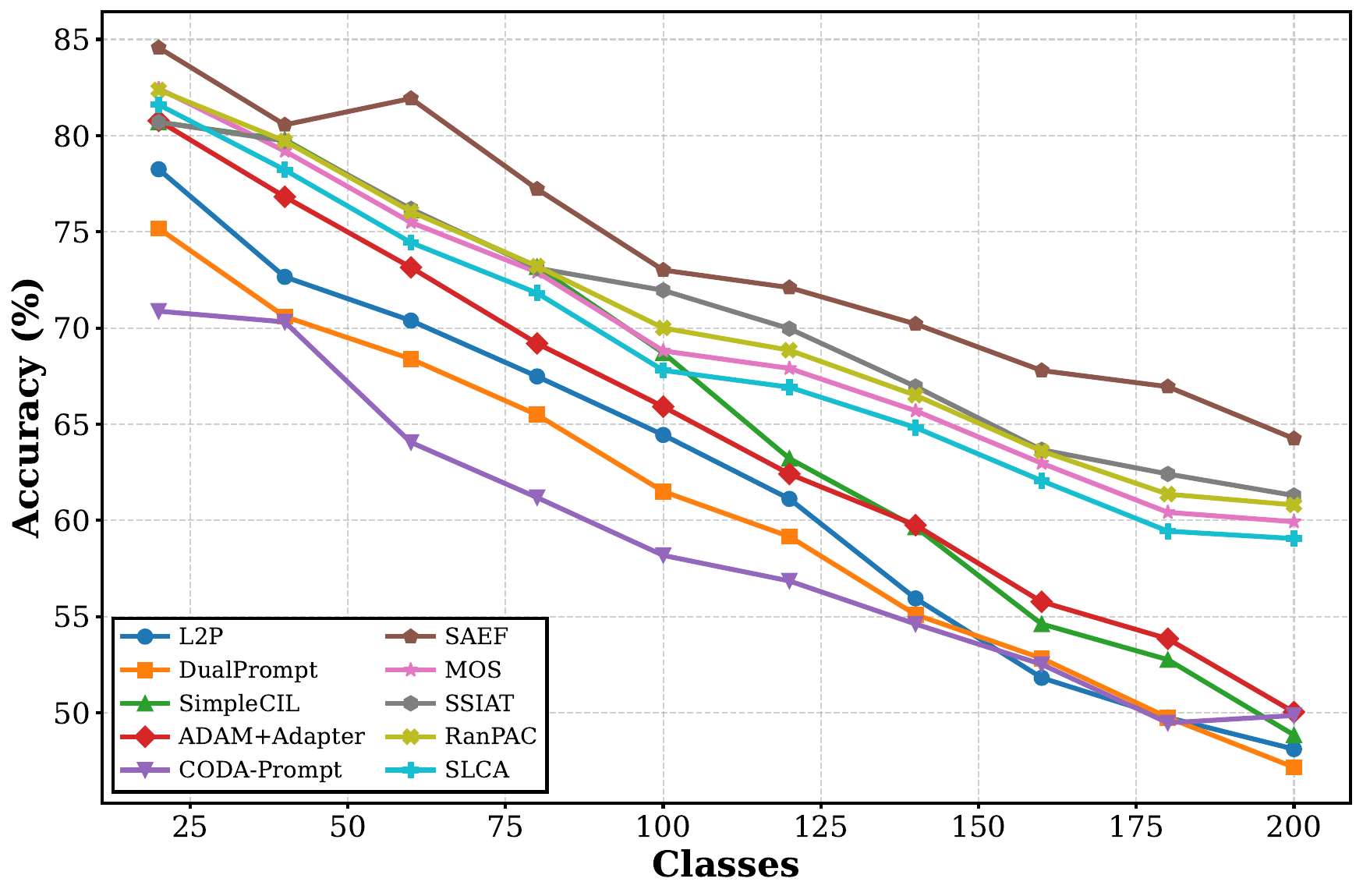}
        \caption{ImageNet-A}
        \label{subfig:appendix_curve_imageneta}
    \end{subfigure}
    \hfill 
    \begin{subfigure}[b]{0.48\textwidth}
        \centering
        \includegraphics[width=\textwidth]{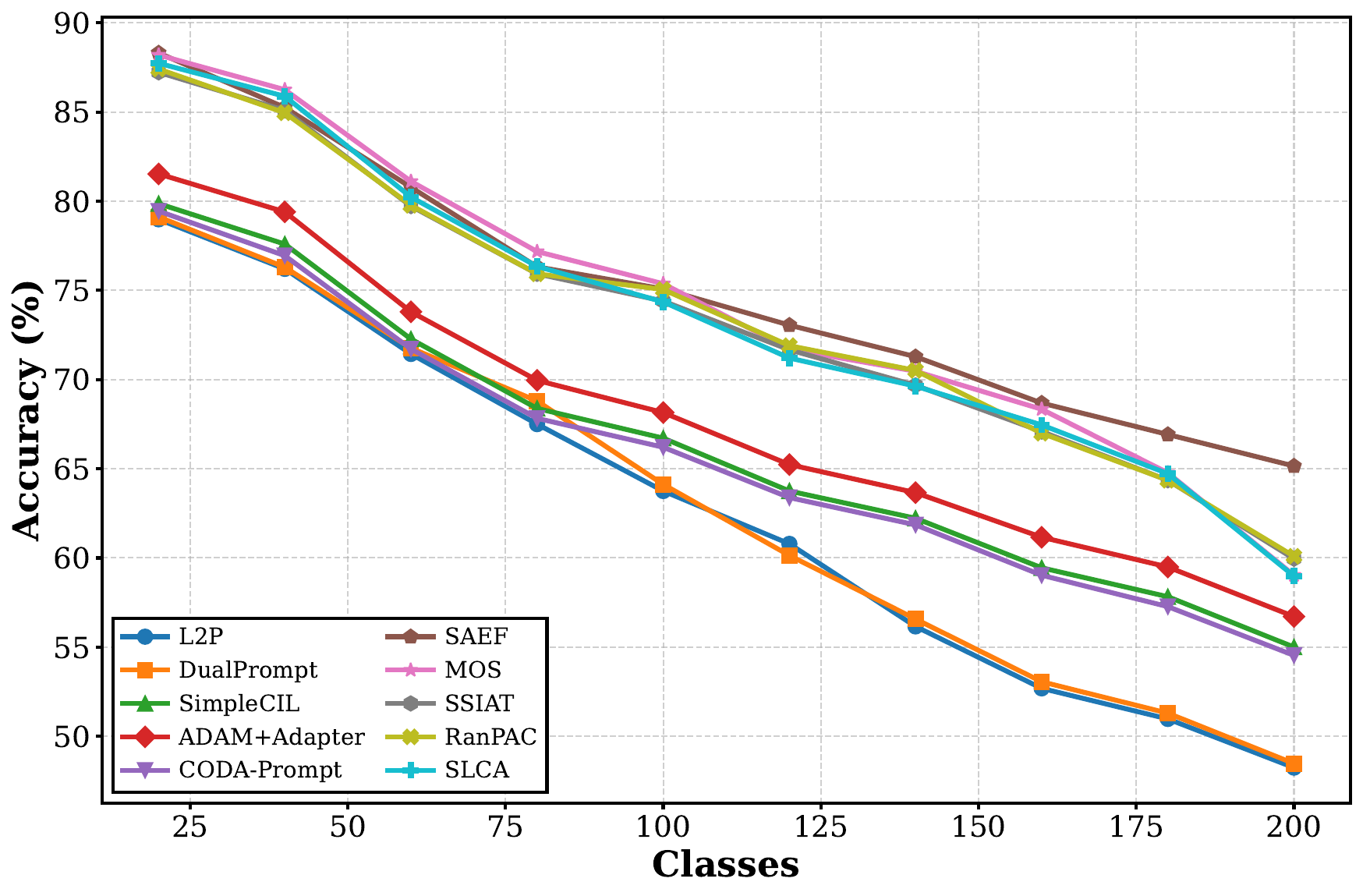}
        \caption{ObjectNet}
        \label{subfig:appendix_curve_objectnet}
    \end{subfigure}

    \caption{
        Incremental accuracy trends of SAEF compared to SOTA CIL methods on four standard benchmarks: CIFAR-100, ImageNet-R, ImageNet-A, and ObjectNet. Across all scenarios, SAEF demonstrates consistently superior performance. This highlights its robust ability to mitigate catastrophic forgetting and maintain a clear advantage as the number of learned classes increases.
    }
    \label{fig:appendix_perf_curves}
\end{figure*}
\begin{figure}[t!]
    \centering
    \includegraphics[width=0.8\linewidth]{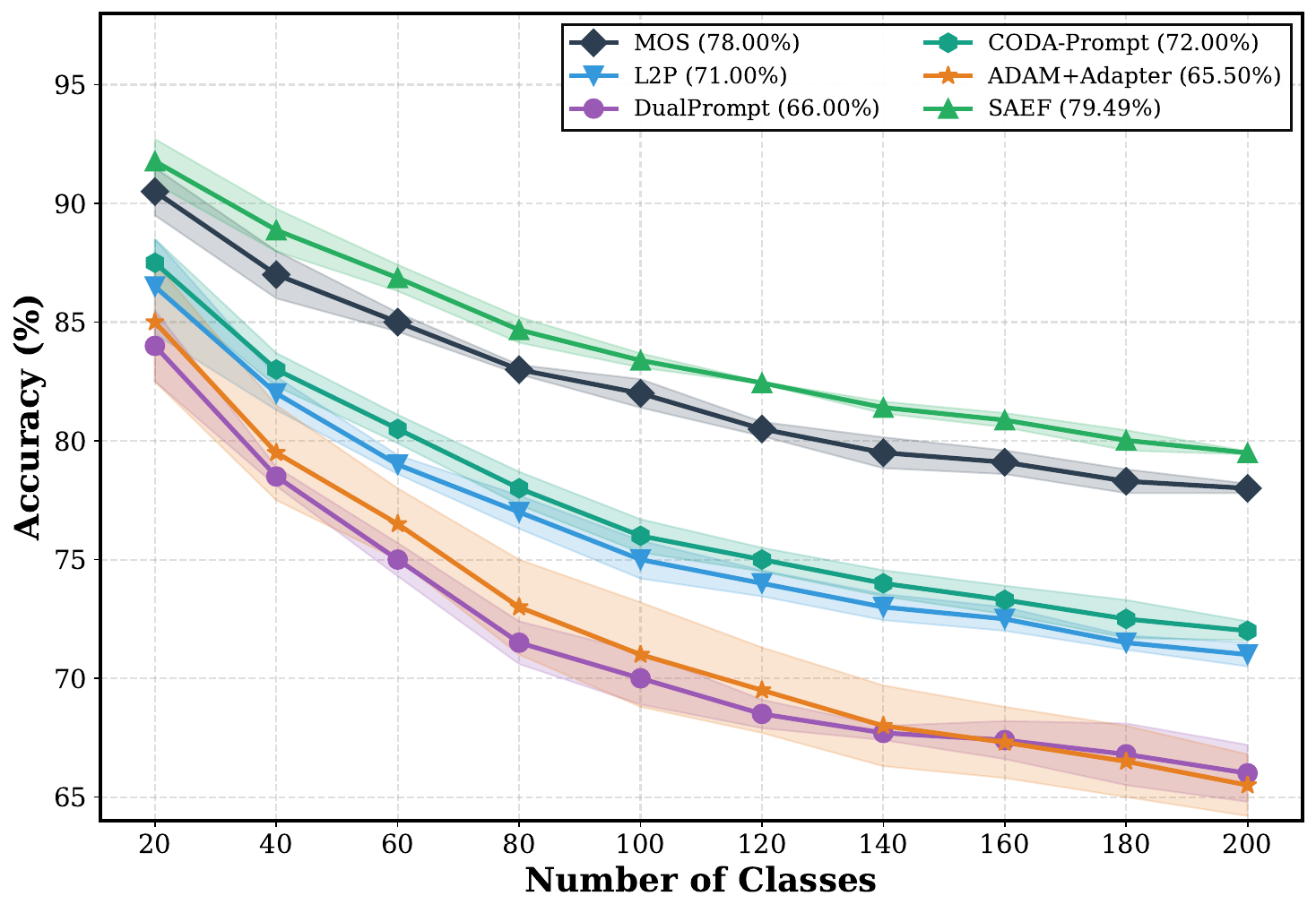} 
    \caption{
        Performance stability across multiple random seeds on ImageNet-R (10-task setting).
        The plot displays the average accuracy ($\bar{\mathcal{A}}$) as the number of learned tasks increases. 
        All results shown are the mean performance calculated over five different random seeds, which vary the order of incoming classes. 
        The consistent and superior performance of SAEF highlights its robustness against variations in the incremental learning sequence.
    }
    \label{fig:multiple_runs_saef}
\end{figure}

\subsection{Algorithmic Details}
\label{app:algo_details}

\paragraph{Offline Hierarchy Construction.}
Algorithm~\ref{alg:saef_construction} details the offline process for organizing learned adapters into the SAEF hierarchy. The \texttt{BuildHierarchy} procedure encapsulates this entire workflow. It begins by computing the semantic and visual prototypes for all tasks (Lines 4-8), following Eq.~\eqref{eq:semantic_prototype} and Eq.~\eqref{eq:visual_prototype}. Based on the semantic prototypes, it automatically determines the optimal number of clusters $K^*$ and partitions the tasks (Lines 9-10). Subsequently, it iterates through each cluster to build a dedicated expert tree using the \texttt{BuildTree} helper function (Lines 13-15). The \texttt{BuildTree} procedure (Lines 21-28) implements the iterative, bottom-up merging process. At each step, it identifies and merges the most visually similar pair of nodes (Lines 23-26), guided by Eq.~\eqref{eq:similarity_metric} and Eq.~\eqref{eq:param_merge}, until a single tree root is formed. Finally, the main procedure merges the roots of all trees into a single global expert (Lines 16-17) as per Eq.~\eqref{eq:global_root_merge}, completing the full hierarchy.

\paragraph{Online Adaptive Inference.}
Algorithm~\ref{alg:saef_inference} details the online prediction mechanism for a given input sample $x$. The \texttt{Inference} procedure first identifies the set of all relevant experts, $\mathcal{E}$ (Lines 4-8). This involves starting with the global expert and then traversing each of the $K^*$ trees to find the optimal path using the recursive helper function \texttt{FindPath} (Line 6). The \texttt{FindPath} procedure (Lines 23-35) implements the entropy-guided traversal, including the early-exit mechanism controlled by the threshold $\tau_e$ (Line 26). Once the set of activated experts is finalized, the algorithm computes the predictive distribution $\mathbf{z}_n$ and corresponding confidence (entropy $H_n$) for each expert (Lines 10-14). Finally, it calculates the normalized, confidence-based fusion weights $w_n$ and computes the final prediction as a weighted average of all activated experts' outputs (Lines 16-20), directly implementing the fusion logic from Eq.~\eqref{eq:fusion_weights} and Eq.~\eqref{eq:final_prediction}.

\section{Theoretical Speedup Calculation}
\label{app:speedup_calc}

In our experiments, we report the theoretical speedup of SAEF's inference process relative to a baseline 'Flat Ensemble' method. The baseline requires querying all $N$ available task-specific adapters. Therefore, its computational complexity is proportional to $N$.

The inference process of SAEF involves two main steps:
\begin{enumerate}
    \item A mandatory forward pass through the global root adapter, $\mathcal{A}_R$.
    \item A parallel search within each of the $K$ conceptual trees. The average number of adapters queried per tree is determined by the average search depth, denoted as $\bar{d}$.
\end{enumerate}

Thus, the total number of adapter queries for SAEF on average is $1 + K \times \bar{d}$.

The theoretical speedup ($S$) is defined as the ratio of the number of queries required by the baseline to the number of queries required by SAEF. It is formulated as:
\begin{equation}
\label{eq:speedup}
S = \frac{\text{Queries}_{\text{Baseline}}}{\text{Queries}_{\text{SAEF}}} = \frac{N}{1 + K \cdot \bar{d}}
\end{equation}
where $N$ is the total number of task-specific adapters (leaf nodes), $K$ is the number of conceptual clusters (trees), and $\bar{d}$ is the average search depth, which is controlled by the entropy threshold $\tau_e$. This formula provides a principled way to estimate the efficiency gains from SAEF's hierarchical structure and its adaptive search mechanism.

\section{Performance Stability Across Multiple Runs}
\label{app:multiple_runs}

In the main paper, our experiments adhere to the standard protocol from Rebuffi et al.~\cite{rebuffi2017icarl}, employing a fixed random seed of 1993 to determine the class order for the incremental tasks. To rigorously assess the robustness of our method against variations in task composition and ordering, this section extends our analysis by repeating the experiments on ImageNet-R with five different random seeds: \{1993, 1994, 1995, 1996, 1997\}.

This procedure generates five distinct result sets for each evaluated method. The mean performance for each method is visualized in Figure~\ref{fig:multiple_runs_saef}. As the figure illustrates, SAEF consistently outperforms all competing methods across the different random seeds. This confirms the stability and general effectiveness of our structured, semantic hierarchy, irrespective of the specific sequence in which classes are introduced.

\section{Detailed Incremental Performance Analysis}
\label{appendix:performance_curves}

To provide a more detailed view of SAEF's performance progression, we present the incremental average accuracy trends against leading CIL methods in Figure~\ref{fig:appendix_perf_curves}. 
These plots, based on a ViT-B/16 backbone pre-trained on ImageNet-1K, track the performance as the number of learned classes increases across our four primary benchmarks.

Across all evaluated settings, the plots clearly show that SAEF not only achieves the highest final accuracy but also maintains a consistent and often widening performance margin throughout the entire learning sequence.
For instance, on the challenging ImageNet-R benchmark (Figure~\ref{subfig:appendix_curve_imagenetr}), the performance of many competing methods degrades more sharply over time, whereas SAEF's trajectory remains stable and superior.
This robust performance underscores the benefit of SAEF's hierarchical knowledge organization, which allows for more effective and targeted knowledge reuse compared to methods that treat adapters as an unstructured collection.
\section{Analysis of Resource Consumption}
\label{app:resource_analysis}

\noindent
The construction of the SAEF expert forest is a highly efficient offline process. For a typical experimental setup with 10 tasks, this entire process completes in under a few seconds on a standard GPU, making the computational overhead negligible. 

The primary overhead of our method is in storage. To construct the hierarchy, SAEF creates a set of new internal-node adapters. This process nearly doubles the total number of adapters that need to be stored. For example, in a typical 10-task experiment, our method creates an additional 8 adapters to build the hierarchical structure on top of the 10 original ones. However, this represents a deliberate design choice. Given the inherently small storage footprint of each lightweight adapter, this modest increase in storage is a highly favorable trade-off for the significant and repeated gains in inference speed achieved by our adaptive search mechanism.

\end{document}